\documentclass[letterpaper, 10 pt, journal, twoside]{IEEEtran}

\IEEEoverridecommandlockouts                              

\usepackage[utf8]{inputenc}
\usepackage{xcolor,colortbl}
\usepackage{graphicx}
\usepackage{subcaption}
\usepackage{lipsum}  
\usepackage{balance}
\usepackage{rotating}
\usepackage{transparent}
\usepackage[font=small]{caption}
\usepackage{arydshln}
\usepackage{amsmath}
\usepackage{array}
\usepackage{tabularx}
\usepackage{makecell}
\usepackage{multirow}
\usepackage{setspace}
\usepackage{booktabs}
\usepackage{amsfonts}
\usepackage{tabu}

\makeatletter
\let\NAT@parse\undefined
\makeatother          
\usepackage{hyperref}
\usepackage{standalone}
\usepackage{algorithm}
\usepackage{algorithmicx}
\usepackage{algpseudocode} 
\usepackage{threeparttable}

\usepackage[nameinlink,capitalise]{cleveref}
\crefformat{footnote}{#2\footnotemark[#1]#3}

\usepackage{tikz}
 \usepackage{textcomp}
\usepackage{pgfplots}
\usetikzlibrary{patterns,positioning,arrows,arrows.meta,calc,shapes,pgfplots.groupplots,fit,backgrounds}
 \usepackage{textcomp}
\usepackage[most]{tcolorbox}

\newcommand\mydownarrow{\rotatebox[origin=c]{-90}{\MVRightarrow}}

\definecolor{pinkish}{RGB}{249,100,255}
\definecolor{cvprblue}{rgb}{0.21,0.49,0.74}
\definecolor{LightCyan}{rgb}{0.849,1,0.969}
\definecolor{arylideyellow}{rgb}{0.91, 0.84, 0.42}
\definecolor{byzantium}{rgb}{0.44, 0.16, 0.39}
\definecolor{soft_pink}{rgb}{1.0, 0.71, 0.76}
\definecolor{lavender}{rgb}{0.8, 0.6, 1.0}
\usepackage{marvosym}

\definecolor{pinkish}{RGB}{249,100,255}

\newcolumntype{g}{>{\columncolor{Gray}}c}
\definecolor{pinkish}{RGB}{249,100,255}

\title{
UADA3D: Unsupervised Adversarial Domain Adaptation for 3D Object Detection with Sparse LiDAR and Large Domain Gaps
}

\author{Maciej K. Wozniak$^{1}$, Mattias Hansson$^{1}$, Marko Thiel$^{2}$, Patric Jensfelt$^{1}$
\thanks{$^{1}$Mattias Hansson, Maciej K. Wozniak, and Patric Jensfelt are with the Division of Robotics, Perception, and Learning, KTH Royal Institute of Technology, Stockholm, Sweden}
\thanks{$^{2}$Marko Thiel is with the Institute for Technical Logistics, Hamburg University of Technology, Germany}

}

\begin{document}

\maketitle

\begin{abstract}
  In this study, we address a gap in existing unsupervised domain adaptation approaches on LiDAR-based 3D object detection, which have predominantly concentrated on adapting between established, high-density autonomous driving datasets. We focus on sparser point clouds, capturing scenarios from different perspectives: not just from vehicles on the road but also from mobile robots on sidewalks, which encounter significantly different environmental conditions and sensor configurations. We introduce Unsupervised Adversarial Domain Adaptation for 3D Object Detection (\textbf{UADA3D}). UADA3D does not depend on pre-trained source models or teacher-student architectures. Instead, it uses an adversarial approach to directly learn domain-invariant features. We demonstrate its efficacy in various adaptation scenarios, showing significant improvements in both self-driving car and mobile robot domains. Our code is open-source and will be available at \url{https://maxiuw.github.io/uda}. 
\end{abstract}

\section{Introduction}
LiDAR-based perception systems are essential for the safe navigation of autonomous vehicles such as self-driving cars or mobile robots. A key challenge is the reliable detection and classification of objects within a vehicle's environment~\cite{wozniak2023towards}. SOTA 3D object detection methods highly depend on quality and diversity of the datasets used for training, but also on how closely these datasets reflect inference conditions. Acquiring and annotating such data remains a significant technical, moral and practical challenge, being both time-consuming and labor-intensive~\cite{wei2024basal}. This presents a major obstacle in development and deployment of 3D object detection models at scale.


A crucial technique to mitigate these challenges is domain adaptation (DA). DA addresses the problem of adapting models trained on a source domain with ample labeled data to a target domain where labels might be scarce (as in semi-supervised DA) or completely unavailable (as in unsupervised DA -- UDA). UDA methods can substantially improve model performance in new, unfamiliar, or changing environments without the need to label new training samples. In the context of autonomous vehicles, discrepancies between source and target domains, often referred to as domain shift or domain gap, can be caused by changes in weather conditions~\cite{li2023domain}, variations in object sizes~\cite{wang2020train}, different sensor setups and deployment environments~\cite{wozniak2023towards,peng2023cl3d} but also due to the transition from simulated to real-world environments~\cite{debortoli2021adversarial}.

\begin{figure}[t]
\centering
 \includegraphics[width=0.75\columnwidth,trim={1cm 0.95cm 0 0},clip]{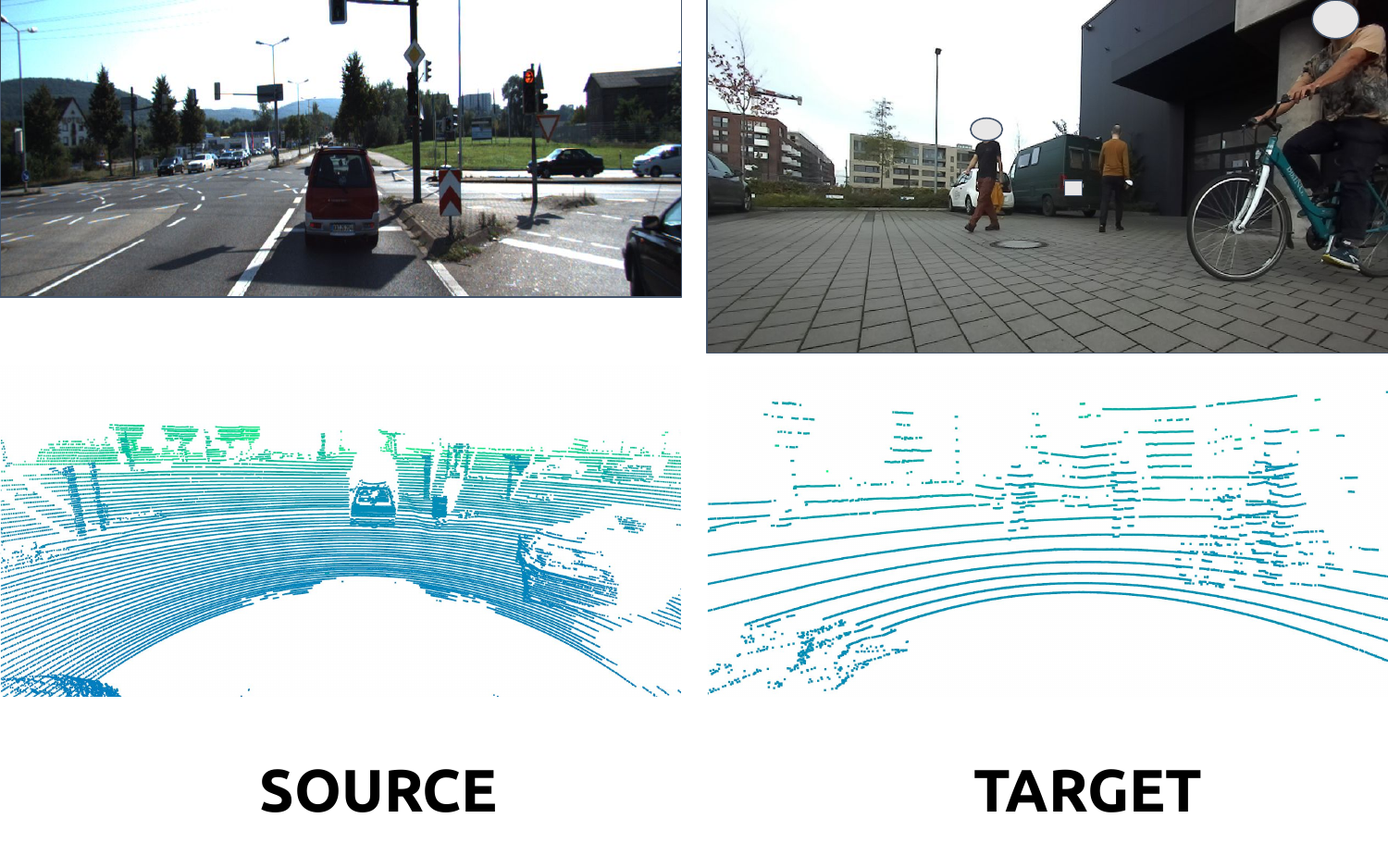}
\caption{Comparison of KITTI (source) and robot data (target). We observe that differences in operating environments, sensor positions, and LiDAR density create a large domain gap. This presents a significant challenge for LiDAR-based 3D object detectors, as well as for the task of domain adaptation.}
\label{fig:domainshift}
\vspace{-0.5cm}
\end{figure}

UDA has received considerable attention in the field of computer vision. However, recent UDA approaches~\cite{luo2021unsupervised,wang2020train,tsai2023ms3d++,chen2023revisiting,li2023gpa} for LiDAR-based 3D object detection primarily focus on automotive applications and corresponding datasets with dense LiDAR data, featuring 128, 64, or 32 layers~\cite{kitti,sun2020waymo,caesar2020nuscenes}. We find a notable research gap when it comes to UDA for 3D object detection models explicitly addressing larger domain shifts than those associated with classical self-driving cars, such as last-mile delivery mobile robots. Those robots operate in an environment sharing many properties with that of self-driving cars, potentially allowing them to benefit from widely available datasets, yet they display significant differences: LiDAR sensors differ in both sensor position and resolution, resulting in sparser point clouds where the ground plane is located much closer to the sensor. Moreover, sidewalk environments are considerably different from roadways. While the same object classes are present, their distribution and relative distances to the sensor are distinctly different than from the car perspective, resulting in a different point density per object, as shown in \cref{fig:domainshift} and further discussed in \cref{sec:Experiments}.

We address UDA in scenarios involving sparse LiDAR and large domain shifts:
1) between widely used automotive datasets, 2) for sim-to-real tasks, and 3) for larger domain shifts using the last-mile delivery robot LAURA~\cite{thiel2023concept} with a 16-layer LiDAR sensor operating on a sidewalk and indoors.


Inspired by 
2D image-based adversarial DA~\cite{ganin2015unsupervised,chen2018domain}, we propose a novel approach for 3D point cloud data: Unsupervised Adversarial Domain Adaptation for 3D Object Detection (\textbf{UADA3D}). Our method uses adversarial adaptation based on class-wise domain discriminators with a gradient reversal layer (GRL) to facilitate the learning of domain-invariant representations. The domain discriminator is trained to maximize its ability to distinguish between the target and source domains, while the model is trained to minimize this ability, resulting in domain-invariant feature learning.

The approach we present offers significant advantages over existing UDA methods for LiDAR-based 3D object detection: UADA3D does not require pre-trained source models, avoids the complexity of teacher-student architectures, and eliminates the inherent uncertainties of pseudo-labels. Instead, it directly learns features that are invariant across the source and target domains. Furthermore, our approach successfully adapts models to multiple object classes simultaneously (e.g. vehicle, pedestrian, and cyclist classes), a capability rarely demonstrated by other SOTA methods~\cite{wei2022lidar,hu2023density,huang2024soap,wang2020train}. Our approach is particularly well-suited for applications with larger domain shifts such as multiple object categories, dissimilar operation environments, and sparse LiDAR data.

Our main contributions are as follows: (i) we introduce UADA3D, an unsupervised adversarial domain adaptation approach for 3D LiDAR-based object detection (\cref{sec:methods});
(ii) we test UADA3D with two widely used object detection models and achieve SOTA performance in various challenging UDA scenarios on autonomous driving and mobile robots datasets (\cref{sec:Experiments} and \cref{sec:results}).

\section{Related Work}
\label{sec:related_work}

\noindent

\textbf{Unsupervised Domain Adaptation for LiDAR-based 3D Object Detection:} Numerous research projects have focused on adapting 3D object detection models between high-resolution LiDAR datasets, ranging from 128 to 32 layers, typically found in self-driving scenarios~\cite{wang2020train,yang2022st3d++,saltori2020sf,tsai2022see,wang2021unsupervised,peng2023cl3d,kong2023conda,tsai2023ms3d++,li2023gpa,yuan2023bi3d,huang2024soap,zhang2024pseudo,zhang2024stal3d}. However, little attention has been given to adapting these models to sparser LiDAR setups, which are common in small robotic platforms~\cite{thiel2023concept}. Peng et al.~\cite{peng2023cl3d} focus specifically on model adaptation between LiDARs with 64 layers and 32 layers, and vice versa. Their results indicate that performance drastically decreases when models trained on 64-layer LiDAR data are adapted to a sparser 32-layer target domain. A similar pattern is observed in ST3D++\cite{yang2022st3d++}, which leverages pseudo-labeling to generate labels for the target domain. MS3D++ \cite{tsai2023ms3d++} leverages multiple pre-trained detectors to achieve more accurate predictions, but again does not focus on very sparse data. Additionally, Cheng et al.~\cite{chen2023revisiting} show that UDA methods does not maintain consistent performance across different object classes, and the majority of the methods adapt only between the Car/Vehicle class~\cite{wei2022lidar,tsai2023ms3d++}. Additionally, they do not focus on sparse LiDAR or different operating environments. 

Recent works, such as DTS~\cite{hu2023density}, which uses a student-teacher architecture with feature-level graph matching, show performance degradation with 16-layer LiDAR, although they do not report quantitative results. LiDAR Distillation (L.D.) \cite{wei2022lidar} focuses exclusively on adapting models to sparser domains by using regression loss to minimize the difference between the BEV-feature maps predicted by student and teacher networks. However, their evaluations are conducted solely on autonomous driving datasets. While they extensively discuss the use of artificially downsampled 16-layer LiDAR data, they do not consider factors such as LiDAR sensor position (e.g., large height shift), domains that differ significantly (e.g., sidewalks versus streets), nor multiple classes. Most of the methods discussed above rely on a student-teacher approach, meaning they require a model pre-trained on the source domain to distill knowledge from the teacher to the student and/or to generate pseudo labels. In contrast, our method employs an adversarial learning approach. Consequently, we do not depend on a pre-trained teacher model to generate, for example, pseudo labels, which are not necessarily high-quality substitutes for ground truth and may even result in decreased model performance, as we later demonstrate in~\cref{sec:results}.

\textbf{Adversarial Domain Adaptation:} Adversarial domain adaptation (ADA) relies on a discriminative network structure that leverages domain discriminators to achieve domain invariance. The GRL~\cite{ganin2015unsupervised} reverses gradients linking the feature extractor $\theta_f$ and the discriminator $\theta_D$. The total objective is to minimize:
$L_{\text{total}} = L_{\text{task}}(\theta_f) - \lambda L_{\text{domain}}(\theta_f, \theta_D)$
where $\lambda$ controls the weight of gradient reversal. Minimizing $L_{\text{task}}$ ensures that the feature extractor learns task-specific features, and minimizing $-\lambda L_{\text{domain}}$ (due to the negative sign) ensures that the feature extractor learns domain-invariant features. This end-to-end training simultaneously employs source and target data, aligning features across domains while minimizing detection loss in the source data.

Following this, Chen et al.~\cite{chen2018domain} introduced a two-stage adaptation method, incorporating image-level and instance-level adaptation. Saito et al.~\cite{saito2019strong} proposed strong and weak adaptation, emphasizing local object feature alignment. He and Zhang~\cite{he2019multi} employed multiple GRLs for global adaptation at different convolutional layers. Xu et al.~\cite{xu2020exploring} introduced categorical regularization between image- and instance-level domain classification, using a regularization loss based on Euclidean distance. Li et al.~\cite{li2023domain} introduced AdvGRL, replacing the constant hyperparameter $\lambda$ with an adaptive $\lambda_{adv}$ to address challenging training samples. ADA has found a lot of use in object detection, classification, or segmentation in image space~\cite{chen2018domain,ganin2015unsupervised,vu2019advent,jaritz2020xmuda}. SRDAN~\cite{zhang2021srdan} or STAL3D~\cite{zhang2024stal3d} use adversarial approaches but their architecture is much different, however, they do not focus on per-class domain prediction, conditional adaptation or analyze different alignment strategies. They focus on adaptation between autonomous driving datasets, not LiDAR sparseness and distinct environments.


\begin{figure*}[ht!]
    \centering
    \includegraphics[trim={.3cm 3.2cm 0 .78cm},clip,width=0.75\textwidth]{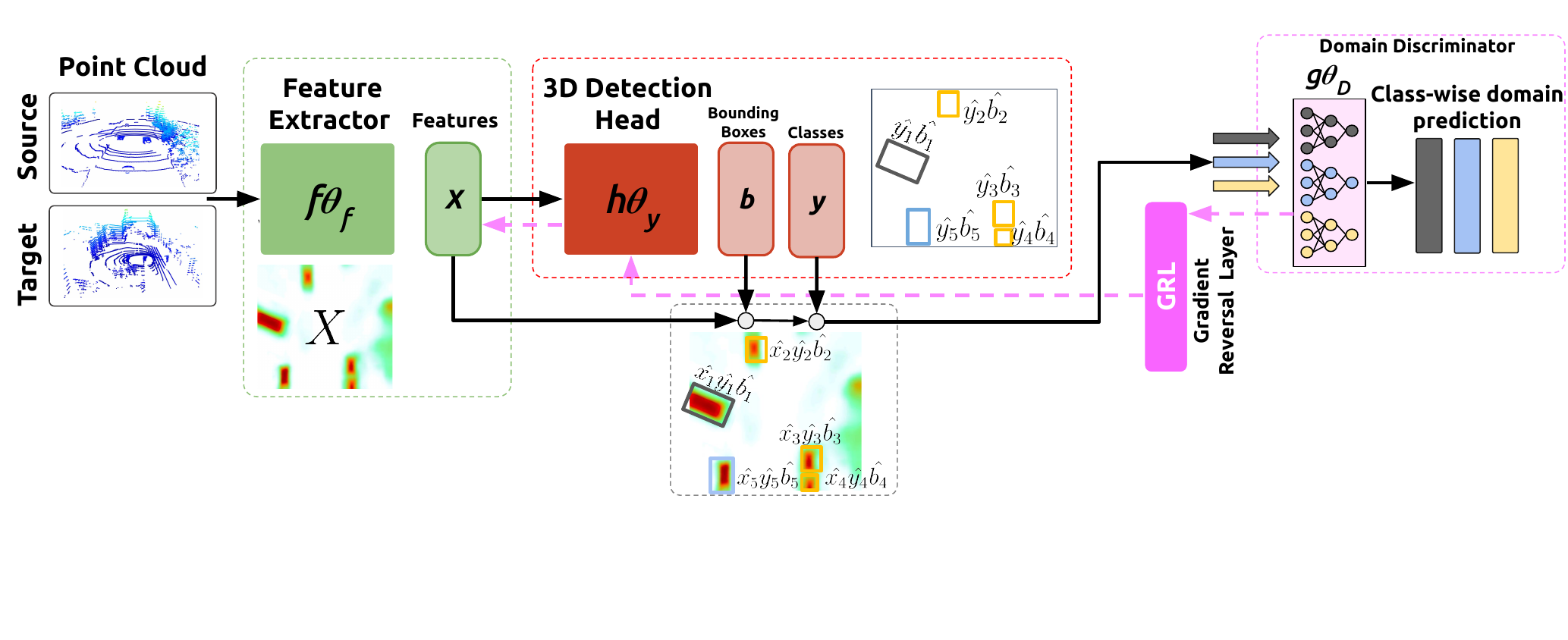}
    \vspace{-0.15cm}
    \caption{An overview of \textbf{UADA3D} (\textbf{black} arrows show forwards, and \textcolor{pinkish}{\textbf{pink}} backward pass). While the primary task of $f_{\theta_f}$ and $h_{\theta_y}$ is 3D object detection, the discriminator $g_{\theta_D}$ aims to classify the domain of each detected instance. Discriminator's loss, reversed by GRL, encourages the detector to learn features that are not only effective for object detection but also invariant across domains.}
    \label{fig:UADA3D_main}
\end{figure*}

Our work concentrates on adversarial domain adaptation in LiDAR-based 3D object detection, distinct from 2D methods. Our approach tackles the unique challenges of spatial data handling, arising from different point cloud densities and operating environments, through data augmentation and method design. We address it particularly with our discriminator utilizing features masked by 3D bounding boxes. Extracting these features from sparser, irregularly spaced point clouds is significantly more challenging than from pixel grids. This often requires transitioning to representations like Bird-Eye-View (BEV) feature space, as illustrated in~\cref{fig:UADA3D_main}. Additionally, working with point clouds involves making predictions based on partially missing, occluded, or incomplete data (please, refer to our website for qualitative results), caused by low angular resolution, potentially resulting in a small number of points per object, even at short distances to the object. In the realm of UDA for segmentation, existing methods employ adversarial approaches
with marginal alignment or non-adversarial methods~\cite{jaritz2020xmuda,liao2023geometry,feng2024open}. Our approach utilizes conditional alignment, yielding superior performance for LiDAR-based 3D object detection (see \cref{sec:ablstudies} for a comparison).

\section{Method}
\label{sec:methods}

\subsection{Problem Formulation}
\label{sec:probabilistic-perspective}

Suppose that \( Q \) is a point cloud, and \( X\) is its feature representation learned by the feature extraction network \( f_{\theta_f} \). The detection head \( h_{\theta_y} \) uses these features to predict \( P(Y|X) \), where  \( Y = (y,b) \) are the category labels \(y\) and bounding boxes \(b\). \( Q \) is sampled from the source domain $\mathcal{D}_s$ and target domain $\mathcal{D}_t$. The objective is to learn generalized weights \( \theta_f \) and \( \theta_y \) between domains such that \( P(Y_s, X_s) \approx P(Y_t, X_t) \). Since \( P(Y, X) = P(Y|X)P(X) \), the domain adaptation task for LiDAR-based object detectors is to align the marginal probability distributions $P(X_s)$ and $P(X_t)$ as well as the conditional probability distributions $P(Y_s|X_s)$ and $P(Y_t|X_t)$. Note that target labels $Y_t$ are not available during training, thus we must use unsupervised domain adaptation.

\subsection{Method Overview}
\label{sec:method-overview}



\begin{algorithm}[b!]
\flushleft
\caption{UADA3D}\label{alg:UDA}
\fontsize{9}{1}\selectfont 
\hspace*{\algorithmicindent} \textbf{Input} Labeled source dataset $\mathcal{D}_s:\{(Q^s,Y^s)\}^{N_s}$, unlabeled target dataset $\mathcal{D}_t:\{(Q^t)\}^{N_t}$ \\
\hspace*{\algorithmicindent} \textbf{Output} Weights: backbone $\theta_f$, detection head $\theta_y$,  discriminators $\theta_D$\\

\begin{algorithmic}[1]
\State $\theta_f,\theta_y,\theta_D \gets \textit{Weight Initialization}$
\For{$Q \in \mathcal{D}_s,\mathcal{D}_t$}
 \State $X \gets f_{\theta_f}(Q)$ 
 \State $\hat{Y} \gets h_{\theta_y}(X)$ \Comment{$\hat{Y}=(\hat{y},\hat{b})$}
\If{source domain}
\State $\theta_y \gets UpdateWeights(\theta_y, \frac{\partial \mathcal{L}_{det}}{\partial \theta_y})$
\State $\theta_f \gets UpdateWeights(\theta_f, \frac{\partial \mathcal{L}_{det}}{\partial \theta_f})$    
\EndIf

 \State $\mathcal{L}_{C} \gets \frac{1}{N}\sum^N \hat{y}_{k,n}\odot(g_{\theta_D,k}(x_n,\hat{b_n})-d)^2$
\State $\theta_D \gets UpdateWeights(\theta_D, \frac{\partial \mathcal{L}_{C}}{\partial \theta_D})$
\State $\theta_y \gets UpdateWeights(\theta_y, -\lambda \frac{\partial \mathcal{L}_{C}}{\partial \theta_y})$
\State $\theta_f \gets UpdateWeights(\theta_f, -\lambda \frac{\partial \mathcal{L}_{C}}{\partial \theta_f})$
\EndFor

\State \Return $\theta_f^*,\theta_y^*$,$\theta_D^*$
\end{algorithmic}

\end{algorithm}





Marginal adaptation, i.e., aligning  $P(X)$, overlooks category and position labels, which can lead to uneven and biased adaptation. This may reduce the target domain's discriminative ability. Aligning \(P(Y | X)\) places direct emphasis on the task-specific outcomes (class labels and bounding boxes) in relation to the features. By focusing on \(P(Y | X)\), we hypothesize that the adaptation process also becomes more robust to variations in feature distributions across domains, concentrating on the essential task of detecting objects. Furthermore, in~\cref{fig:distandsize,fig:ptscls}, we show that the distribution of points in different categories per object varies significantly between the datasets due to LiDAR density, position and operating environment, while the vehicle size slightly varies, depending on the dataset country of origin. While we chose to use conditional alignment, we compare our method with different alignment strategies: marginal and joint distribution alignment in~\cref{sec:ablstudies}.

\cref{fig:UADA3D_main} provides a schematic overview of our method UADA3D. In each iteration, a batch of samples $Q$ from source $\mathcal{D}_s$ and target $\mathcal{D}_t$ domain is fed to the feature extractor $f_{\theta_f}$. In each batch we sample from source and target training splits. The number of samples from the target and source data differ in each batch since the probability of drawing the sample is proportional to the dataset size and the model sees each sample once per epoch. Next, for each sample, features are extracted, and fed to the detection head $h_{\theta_y}$ that predicts 3D bounding boxes (lines 3-4 in \cref{alg:UDA}). The object detection loss is calculated only for the labeled samples from source domain (lines 6-7). The probability distribution alignment branch uses the domain discriminator $g_{\theta_D}$ (line 9) to predict from which domain samples came from, based on the extracted features $X$ and predicted labels $\hat{Y}$. The domain loss $\mathcal{L_C}$ is calculated for all samples (line 9). Next, $\mathcal{L_C}$ is backpropagated through the discriminators (line 10) and through the gradient reversal layer (GRL) with the coefficient $\lambda$, that reverses the gradient during backpropagation, to the detection head and feature extractor (lines 11-12). This adversarial training scheme works towards creating domain invariant features. Thus, our network learns how to extract features that will be domain invariant but also how to provide accurate predictions. Therefore, we seek for the optimal parameters $\theta_f^*$, $\theta_y^*$, and $\theta_D^*$, that satisfy:

\begin{equation}
\begin{array}{cc}
     \theta_D^* = \arg\!\min\limits_{\theta_D}\mathcal{L}_C   \\
     (\theta_f^*,\theta_y^*) = \arg\!\min\limits_{\theta_f,\theta_y}\mathcal{L}_{det} - \lambda \mathcal{L}_C 
\end{array}
\end{equation}

\noindent
where $\mathcal{L}_{det}$ is the detection loss (described in \cite{yin2021center,zhang2022not}) calculated only for the labeled source domain, $\mathcal{L}_C$ is the domain loss calculated for both source and target domains. 

Our approach builds on the method by Ganin and Lempitsky~\cite{ganin2015unsupervised}. However, what they proposed is more similar to the marginal adaptation approach. They do not mask the features with the predicted bounding boxes but predict the domain based on the global features. We introduce feature masking and a class-wise domain discriminator. Additionally, we use prediction confidence to weight discriminator loss, such that low confidence predictions do not influence the performance. These novel aspects and 3D application distinguish us from the approach proposed by Ganin and Lempitsky.

\subsection{Feature Masking}
\label{sec:method-masking}

\textit{Feature masking} plays a crucial role in predicting the domain based on specific object features. Masking enables the model to focus solely on the features corresponding to each instance, thus enhancing the relevance and accuracy of the domain prediction. In \cref{fig:UADA3D_main}, we show how features extracted from a point cloud $Q$ are masked and used for domain prediction. The input to the class-wise domain discriminators $g_{\theta_{D,k}}$ is $(x,\hat{b})$, where $x$ are masked features, $\hat{b}$ are predcited bounding boxes, and $(a,b)$ denotes a concatenation. To obtain masked features $x$, we mask the feature map $X=f_{\theta f}(Q)$ with each predicted bounding box $\hat{b}_n$ creating corresponding masked features $x_n$. Finally, we concatenate $x_n$ with the bounding box $\hat{b}_n$ and feed to the corresponding class-wise discriminator $g_{\theta_{D,k}}$. We further discuss the importance of feature masking in~\cref{sec:ablstudies}. 


\subsection{Conditional Distribution Alignment}
\label{sec:method-conditional}
The conditional distribution alignment module, shown in \cref{fig:UADA3D_main}, has the task of reducing the discrepancy between the conditional distribution $P(Y_s|X_s)$ of the source and $P(Y_t|X_t)$ of the target. As we highlighted in~\cref{sec:method-overview}, show in~\cref{fig:ptscls} and later discuss in~\cref{sec:setup-datasets}, we can see a large difference between how objects from each category appear in different domains. Thus, instead of having one discriminator, we use $K=3$ class-wise domain discriminators $g_{\theta_D,k}$, corresponding to vehicle, pedestrian, and cyclist classes. The conditional distribution alignment module is trained using the least-squares loss function:

\begin{equation}
    \mathcal{L}_{C} = \frac{1}{N}\sum_{n=1}^N\hat{y}_{k,n} \odot (g_{\theta_D,k}(x_n,\hat{b}_n)-d)^2
   \label{eq:loss_cond}
\end{equation}

\noindent
where $N$ is the number of labels, $\hat{y}_{k,n}$ corresponding class confidence of instance $n$ and $\odot$ is element-wise multiplication. The loss is backpropagated to the discriminators (line 9 in \Cref{alg:UDA}). Next, $\mathcal{L}_C$ is backpropagated through GRL to the detection head $h_{\theta_y}$ and the feature extractor $f_{\theta_f}$. 
\subsection{Data Augmentation}
\label{sec:DS}

Differences between LiDAR domains include different densities and object sizes. We use downsampling, a commonly used augmentation approach since the domain gap can be partially remedied by reducing LiDAR layers of the source data to $16$ or $32$ to better match the target domain LiDAR data as highlighted in~\cite{fang2023lidar}. LiDAR-CS~\cite{fang2023lidar}, Waymo~\cite{sun2020waymo}, and nuScenes~\cite{caesar2020nuscenes} datasets contain vehicle sizes that correspond to the large vehicle sizes found in the USA, while our robot and KITTI~\cite{kitti} are collected in Europe where vehicles are smaller. Random object scaling (ROS)~\cite{yang2022st3d++} is applied to source domain to address this vehicle size bias. While previous UDA methods on LiDAR-based 3D object detection often apply domain adaptation only to a single object category, we consider it a multiclass problem. Thus, we chose ROS with different scaling intervals for the three classes. The interested reader can find a more detailed analysis of the models' performance with ROS and downsampling on the project's webpage. Finally, points of both domains were shifted in vertical direction so that $z=0$ at the position of the ground plane in sensor coordinates. 

\section{Experiments Setup}
\label{sec:Experiments}


We compare performance of IA-SSD~\cite{zhang2022not} and Centerpoint~\cite{yin2021center} on large number of UDA scenarios using our method, UADA3D, against that of other SOTA unsupervised domain adaptation approaches (ST3D++~\cite{yang2022st3d++}, DTS~\cite{hu2023density}, L.D.~\cite{wei2022lidar}, and MS3D++~\cite{tsai2023ms3d++}). We chose IA-SSD due to its efficiency, speed and low-memory requirements as a potential good fit for robotics application. Centerpoint is more robust detector achieving higher performance, however, with larger computational cost. In IA-SSD, the adaptive discriminator network is made of fully-connected layers that operate on down-sampled point features. In Centerpoint, the discriminator makes use of 2D convolutions with inputs from BEV feature maps. The most prominent distinction between the two networks is the point-based and view-based representations, which are handled by MLPs and 2D convolutions respectively. The ability to adapt both of these models between the domains shows the modularity of our solution and adaptivity to different methods. With UADA3D, Centerpoint is trained for 40 and IA-SSD for 80 epochs using one-cycle Adam with a learning rate 0.003 and 0.01 respectively. The GRL-coefficient was set to a fixed value of 0.1. For all the SOTA methods, we train the source-only and oracle models with 40 (Centerpoint) and 80 (IA-SSD) epochs, then we adapt them for 40 and 80 accordingly. 


     

\subsection{Datasets}
\label{sec:setup-datasets}

\begin{figure}[t]

     \begin{subfigure}[b!]{0.24\textwidth}
         \centering    
         \begin{tikzpicture}
\begin{axis} [ybar = .05cm,
    bar width = 4.25pt, 
    width=1.5\textwidth,
    height=0.8\textwidth,
    tick pos=left,
   xmin = 0, 
    xmax = 70, 
    ymin=0,
    ymax=20,
    enlarge x limits = {abs=0.5},
    tick label style={font=\normalsize},
    legend style={font=\footnotesize},
    label style={font=\small},
    y label style={at={(axis description cs:0.17,.5)},anchor=south},
    x label style={at={(axis description cs:0.97,0.24)},anchor=north},
     legend style={legend columns=-1,at={(0.5,1.32)},anchor=north},
    legend image code/.code={
        \draw [#1] (0cm,-0.1cm) rectangle (0.1cm,0.15cm);},
    bar shift = 0.0,
    xlabel = (m),
    ylabel = Percentage (\%)
]
\addplot [fill=orange!60,opacity=0.6] coordinates{
(1.25, 0.05) (3.75, 3.53) (6.25, 3.73) (8.75, 4.02) (11.25, 4.35) (13.75, 4.95) (16.25, 5.50) (18.75, 5.78) (21.25, 6.23) (23.75, 6.40) (26.25, 5.92) (28.75, 5.58) (31.25, 5.38) (33.75, 5.03) (36.25, 4.88) (38.75, 4.52) (41.25, 4.25) (43.75, 3.87) (46.25, 3.46) (48.75, 2.96) (51.25, 2.68) (53.75, 2.33) (56.25, 2.05) (58.75, 1.64) (61.25, 0.79) (63.75, 0.10) (66.25, 0.00)
};

\addplot [fill=teal!60,opacity=0.6] coordinates{
(1.25, 0.03) (3.75, 0.51) (6.25, 2.66) (8.75, 6.64) (11.25, 10.56) (13.75, 12.90) (16.25, 15.48) (18.75, 15.01) (21.25, 11.69) (23.75, 8.81) (26.25, 6.40) (28.75, 4.25) (31.25, 2.56) (33.75, 1.21) (36.25, 0.68) (38.75, 0.22) (41.25, 0.14) (43.75, 0.13) (46.25, 0.04) (48.75, 0.02) (51.25, 0.03) (53.75, 0.01) (56.25, 0.00) (58.75, 0.00) (61.25, 0.00) (63.75, 0.00) (66.25, 0.00)
};

\addplot [fill=purple!60,opacity=0.6] coordinates{
(1.25, 0.13) (3.75, 0.83) (6.25, 2.37) (8.75, 4.59) (11.25, 6.98) (13.75, 7.75) (16.25, 9.67) (18.75, 9.30) (21.25, 10.62) (23.75, 9.41) (26.25, 8.88) (28.75, 7.39) (31.25, 5.82) (33.75, 4.72) (36.25, 3.73) (38.75, 2.35) (41.25, 1.88) (43.75, 1.24) (46.25, 0.67) (48.75, 0.60) (51.25, 0.39) (53.75, 0.31) (56.25, 0.21) (58.75, 0.13) (61.25, 0.01) (63.75, 0.00) (66.25, 0.00)
};

\legend {Vehicle, Pedestrian, Cyclist};
 
\end{axis} 
\end{tikzpicture}

         \vspace{-0.6cm}
         \caption{\textbf{Distance} to objects \textit{LiDAR-CS}}
         \label{fig:distlidar}
     \end{subfigure}
     \hfill
     \begin{subfigure}[b!]{0.24\textwidth}
         \centering
         \begin{tikzpicture}

\begin{axis} [ybar = .05cm,
    bar width = 4.25pt, 
    width=1.5\textwidth,
    height=0.8\textwidth,
    tick pos=left,
    xmin = 0, 
    xmax = 17, 
    ymin=0,
    ymax=60,
    enlarge x limits = {abs=0.5},
    tick label style={font=\normalsize},
    legend style={font=\footnotesize},
    label style={font=\small},
    y label style={at={(axis description cs:0.21,.5)},anchor=south},
    x label style={at={(axis description cs:0.97,0.24)},anchor=north},
     legend style={legend columns=-1,at={(0.5,1.32)},anchor=north},
    legend image code/.code={
        \draw [#1] (0cm,-0.1cm) rectangle (0.1cm,0.15cm);},
    bar shift = 0.0,
    xlabel = (m),
]
\addplot [fill=orange!60,opacity=0.6] coordinates{
(0.25, 0.00) (0.75, 0.00) (1.25, 0.00) (1.75, 0.00) (2.25, 0.00) (2.75, 0.00) (3.25, 0.00) (3.75, 0.00) (4.25, 0.00) (4.75, 0.00) (5.25, 0.00) (5.75, 0.00) (6.25, 0.00) (6.75, 0.00) (7.25, 0.00) (7.75, 0.00) (8.25, 0.00) (8.75, 0.00) (9.25, 0.00) (9.75, 0.12) (10.25, 0.73) (10.75, 15.35) (11.25, 41.41) (11.75, 29.60) (12.25, 3.53) (12.75, 2.19) (13.25, 1.71) (13.75, 1.22) (14.25, 1.22)
};

\addplot [fill=teal!60,opacity=0.6] coordinates{
(0.25, 0.00) (0.75, 0.81) (1.25, 6.39) (1.75, 10.65) (2.25, 5.58) (2.75, 8.16) (3.25, 6.09) (3.75, 7.10) (4.25, 6.09) (4.75, 7.76) (5.25, 6.29) (5.75, 3.35) (6.25, 3.60) (6.75, 4.67) (7.25, 4.11) (7.75, 3.04) (8.25, 3.80) (8.75, 1.93) (9.25, 1.67) (9.75, 1.32) (10.25, 1.27) (10.75, 0.81) (11.25, 1.57) (11.75, 0.25) (12.25, 0.51) (12.75, 0.46) (13.25, 0.30) (13.75, 0.46) (14.25, 0.25)
};

\addplot [fill=purple!60,opacity=0.6] coordinates{
(0.25, 0.00) (0.75, 0.00) (1.25, 0.25) (1.75, 8.25) (2.25, 5.25) (2.75, 5.25) (3.25, 4.25) (3.75, 4.50) (4.25, 3.50) (4.75, 5.00) (5.25, 5.50) (5.75, 5.25) (6.25, 5.75) (6.75, 6.75) (7.25, 9.50) (7.75, 7.00) (8.25, 8.25) (8.75, 9.25) (9.25, 4.50) (9.75, 2.00) (10.25, 0.00) (10.75, 0.00) (11.25, 0.00) (11.75, 0.00) (12.25, 0.00) (12.75, 0.00) (13.25, 0.00) (13.75, 0.00) (14.25, 0.00)
};

\legend {Vehicle, Pedestrian, Cyclist};
 
\end{axis} 
\end{tikzpicture}
         \vspace{-0.6cm}
         \caption{\textbf{Distance} to objects \textit{robot data}}
         \label{fig:distlaura}
     \end{subfigure}

        \caption{Comparison of objects in LiDAR-CS and robot datasets.}
        \label{fig:distandsize}
        \centering
\vspace{0.15cm}
\begin{minipage}[t]{.9\columnwidth}
        \centering
         \begin{tikzpicture}

\begin{axis} [ybar = .1cm,
    height=5.5cm,
    width=17cm,
    bar width = 11pt,
    tick pos=left,
    xmin = 0, 
    xmax = 2, 
    ymin=0,
    ymax=1000,
    enlarge x limits = {abs=0.5},
    xtick = data,
    xticklabels = {Vehicle, Pedestrian, Cyclist},
    x tick label style={font=\huge},
    y tick label style={font=\huge},
    legend style={font=\huge},
    x label style={font=\huge},
    legend style={legend columns=-1,at={(0.5,1.33)},anchor=north},
    legend image code/.code={
        \draw [#1] (0.5cm,-0.1cm) rectangle (0.1cm,0.15cm);},
    nodes near coords style={font=\tiny},
]

\addplot coordinates {(0,919) (1,84) (2,143)};
\addplot coordinates {(0,767) (1,201) (2,221)}; 
\addplot coordinates {(0,833) (1,122) (2,199)}; 
\addplot coordinates {(0,611) (1,101) (2,153)};
\addplot coordinates {(0,669) (1,20) (2,53)};
\addplot coordinates {(0,125) (1,13) (2,21)}; 
\addplot [fill=teal!60,opacity=0.6] coordinates {(0,133) (1,456) (2,527)}; 

\legend {CS64, KITTI, Waymo, nuScenes, CS32, CS16, Robot};
 
\end{axis} 
\end{tikzpicture}
         \vspace{-0.6cm}
         \caption{Average number of points in an object per class.}
         \label{fig:ptscls}
\end{minipage}
\end{figure}

We use five datasets: LiDAR-CS~\cite{fang2023lidar} that provides multiple LiDAR resolutions (we use VLD-64 ``\textit{CS64}'', VLD-32 ``\textit{CS32}'', and VLP-16 ``\textit{CS16}''), KITTI~\cite{kitti} ``\textit{K}'' (with HDL-64E), Waymo~\cite{sun2020waymo} ``\textit{W}'' (HDL-64E), nuScenes~\cite{caesar2020nuscenes} ``\textit{N}'' (VLD-32)  and data collected with the last mile delivery robot ``\textit{R}'' (with VLP-16) in different locations in Europe. 

First, we focus on UDA between autonomous driving datasets, from denser to sparser LiDARs, since this is where the other SOTA methods perform the worst. Even though these sensors operate similarly, differences arise from the point-cloud density or LiDARs azimuth and elevation angles. Second, we adapt to our robot data, which is particularly challenging due to different LiDAR positions and densities on the mobile robot, as well as the operating environment, sidewalk vs. street (the dataset will be publicly available). Finally, we adapt from sparser to denser domains and other popular UDA benchmarks used by SOTA.




In \cref{fig:distandsize} we can see how big the gap between the self-driving datasets and robot data is, especially in distance to the objects (nuScenes and Waymo show a similar distribution to LiDAR-CS and KITTI). The distance to the objects encountered in the robot data, seen in \Cref{fig:distlaura}, shows that the dataset only contains objects closer than $15$m, with a majority of pedestrians and cyclists closer than $10$m and all vehicles in the $10-15$m range, since the robot drives on a sidewalk. In contrast, in the self-driving cars dataset \cref{fig:distlidar}, objects are between $20-30$m away. Consequently, the average number of points per object (\cref{fig:ptscls}) is very different between the datasets. This is causing the instances to significantly differ between these domains. While the different vehicle sizes are often addressed in 3D domain adaptation approaches~\cite{yang2022st3d++}, to our knowledge, the differing number of points per object has not been examined by other UDA works before. KITTI dataset is limited by having labels only in the front camera FOV, while other datasets feature labels in a 360-degree FOV. This presents a significant challenge for domain adaptation.

\definecolor{LightCyan}{rgb}{0.849,1,0.969}
\newcolumntype{a}{c}
\newcolumntype{b}{>{\columncolor{Gray}}c}
\definecolor{Gray}{gray}{0.85}

\begin{table*}[]
\Large\selectfont
\centering
\caption{Comparison of performance of different adaptation methods on source to target (S$\rightarrow$T) domain adaptation tasks. We report $mAP_{3D}$ and $mAP_{BEV}$over Vehicle, Pedestrian and Cyclist. The best score is \textbf{bold} and the second best is \underline{underline}.}
\vspace{-0.2cm}
\label{tab:adaptationMain}
\resizebox{0.67\textwidth}{!}{

\begin{tabular}{b|c|cccccc}
\hline
&                           & \multicolumn{6}{c}{Models}                                                                                                                                                                                                                        \\ \cline{3-8} 
&                           & \multicolumn{3}{c|}{IA-SSD~\cite{zhang2022not}}                                                               & \multicolumn{3}{c}{Centerpoint~\cite{yin2021center}}                                   \\ \cline{3-8} 
\multirow{-3}{*}{S\MVRightarrow T}                             & \multirow{-3}{*}{Methods} & \multicolumn{1}{c|}{3D/BEV}           & \multicolumn{1}{c|}{Change}                                & \multicolumn{1}{a|}{Closed Gap} & \multicolumn{1}{c|}{3D/BEV}           & \multicolumn{1}{c|}{Change}                                & Closed Gap \\ \hline


& Source Only               & \multicolumn{1}{c|}{1.5 / 4.5}          & \multicolumn{1}{c|}{-/-}                                     & \multicolumn{1}{a|}{-/-}           & \multicolumn{1}{c|}{15.96 / 17.87}          & \multicolumn{1}{c|}{-/-}                                     &   -/-        \\ \cline{2-8} 

& ST3D++~\cite{yang2022st3d++}                   & \multicolumn{1}{c|}{11.94 / 13.15}          & \multicolumn{1}{c|}{{\color[HTML]{004000} 10.45 / 8.65}}          & {\color[HTML]{004000} 23.38 \% / 20.09 \%}           &

\multicolumn{1}{c|}{17.66 / 22.83}          & \multicolumn{1}{c|}{{\color[HTML]{004000}    1.71} / {\color[HTML]{004000}  4.97}} & {\color[HTML]{004000}    4.6 \%} / {\color[HTML]{004000}   13.45 \%}           \\ \cline{2-8} 

& DTS~\cite{hu2023density}                     & \multicolumn{1}{c|}{15.82 / 17.40}          & \multicolumn{1}{c|}{{\color[HTML]{004000} 14.32 / 12.90}}          & \multicolumn{1}{a|}{{\color[HTML]{004000} 32.03 \% / 29.97 \%}} 

& \multicolumn{1}{c|}{16.24 / 17.89}           & \multicolumn{1}{c|}{{\color[HTML]{004000}    0.28 / 0.02}}        & {\color[HTML]{004000}    0.78 \% /    0.05 \%}            \\ \cline{2-8} 

\multirow{-2}{*}{W} & L.D.~\cite{wei2022lidar}                      & \multicolumn{1}{c|}{\underline{17.57} / \textbf{19.33}}          & \multicolumn{1}{c|}{{\color[HTML]{004000} 16.07 / \textbf{14.83}}}          & \multicolumn{1}{a|}{{\color[HTML]{004000} 35.95 \% / \textbf{34.46} \%}} 
& \multicolumn{1}{c|}{\underline{22.21} / \underline{27.98}}               & \multicolumn{1}{c|}{{\color[HTML]{004000}    6.26} / {\color[HTML]{004000}    10.12}}          & {\color[HTML]{004000}    16.96 \%} / {\color[HTML]{004000}    27.41 \%} \\ \cline{2-8}

\multirow{-2}{*}{\mydownarrow}& MS3D++~\cite{tsai2023ms3d++}                       & \multicolumn{1}{c|}{15.10 / 16.61}          & \multicolumn{1}{c|}{{\color[HTML]{004000} 13.60 / 12.11}}          & \multicolumn{1}{a|}{{\color[HTML]{004000} 30.43 \% / 28.15 \%}} 
& \multicolumn{1}{c|}{19.71/ 21.69}               & \multicolumn{1}{c|}{{\color[HTML]{004000}    3.75} / {\color[HTML]{004000} 3.82}}          & {{\color[HTML]{004000}    10.17 \%} / {\color[HTML]{004000} 10.35 \%}} \\ \cline{2-8}

\multirow{-2}{*}{N}& \textbf{UADA3D (ours)}    & \multicolumn{1}{c|}{\textbf{18.33} / \underline{18.55}} & \multicolumn{1}{c|}{\color[HTML]{004000} \textbf{16.83} / 14.05} & \multicolumn{1}{c|}{\color[HTML]{004000} \textbf{37.66} \% / 32.66 \%} & \multicolumn{1}{c|}{\textbf{26.89 / 30.67}} & \multicolumn{1}{c|}{{\color[HTML]{004000}  \textbf{10.94 }/ \textbf{12.80}}} & {\color[HTML]{004000} \textbf{29.65} \%} / {\color[HTML]{004000} \textbf{34.68 }\%} \\ \cline{2-8} 

 & Oracle                    & \multicolumn{1}{c|}{46.20 / 47.53}          & \multicolumn{1}{c|}{-/-}                                      & \multicolumn{1}{c|}{-/-}           & \multicolumn{1}{c|}{52.84 / 54.77}          & \multicolumn{1}{c|}{-/-}                                      & -/-            \\ \hline \hline



& Source Only               & \multicolumn{1}{c|}{20.11 / 25.26}          & \multicolumn{1}{c|}{-/-}                                      & \multicolumn{1}{c|}{-/-}           & \multicolumn{1}{c|}{22.33 / 38.08}          & \multicolumn{1}{c|}{-/-} & -/-                     \\ \cline{2-8}

& ST3D++~\cite{yang2022st3d++}                    & \multicolumn{1}{c|}{30.11 / 40.09}          & \multicolumn{1}{c|}{{\color[HTML]{004000}    10.00} / {\color[HTML]{004000}    14.83}}          & \multicolumn{1}{c|}{{\color[HTML]{004000}    36.09 \%} / {\color[HTML]{004000}    48.63}}           & \multicolumn{1}{c|}{\textbf{43.90}/50.42}           & \multicolumn{1}{c|}{{\color[HTML]{004000}    \textbf{21.57}} / {\color[HTML]{004000}    12.34}}          & {\color[HTML]{004000}    \textbf{73.99} \%} / {\color[HTML]{004000}    57.40 \%}            \\ \cline{2-8} 

& DTS~\cite{hu2023density}                       & \multicolumn{1}{c|}{\underline{33.41}/\underline{42.62}}               & \multicolumn{1}{c|}{{\color[HTML]{004000}    13.29} / {\color[HTML]{004000}    17.36}}                                        & \multicolumn{1}{c|}{{\color[HTML]{004000}    47.98 \%} / {\color[HTML]{004000}    56.92 \%}}            & \multicolumn{1}{c|}{39.19 / 47.81}          & \multicolumn{1}{c|}{{\color[HTML]{004000}    16.86} / {\color[HTML]{004000}    9.73}}           & {\color[HTML]{004000}    57.86 \%} / {\color[HTML]{004000}    45.26 \%}            \\ \cline{2-8} 

\multirow{-2}{*}{CS64} & L.D.~\cite{wei2022lidar}                        & \multicolumn{1}{c|}{29.81 / 35.35}               & {\color[HTML]{004000}     9.70} / {\color[HTML]{004000}     10.10}                         & \multicolumn{1}{|c}{{\color[HTML]{004000}     34.99 \%} / {\color[HTML]{004000}     33.11 \%}}  & \multicolumn{1}{|c|}{42.66/ \textbf{54.07}}               & \multicolumn{1}{c|}{{\color[HTML]{004000}    20.33} / {\color[HTML]{004000}    \textbf{15.99}}}          & {\color[HTML]{004000}    69.76 \%} / {\color[HTML]{004000}    \textbf{74.38} \%} \\ \cline{2-8}

\multirow{-2}{*}{\mydownarrow}& MS3D++~\cite{tsai2023ms3d++}                        & \multicolumn{1}{c|}{32.79 / 37.21}               & \multicolumn{1}{c|}{{\color[HTML]{004000}    12.68} / {\color[HTML]{004000}    11.95}}                          & \multicolumn{1}{c|}{{\color[HTML]{004000}    45.77 \%} / {\color[HTML]{004000}    39.17 \%}}  & \multicolumn{1}{|c|}{30.98 / 38.53}               & \multicolumn{1}{c|}{{\color[HTML]{004000}    8.65} / {\color[HTML]{004000}    0.45}}          & {\color[HTML]{004000}    29.68 \%} /  {\color[HTML]{004000}    2.08 \%} \\ \cline{2-8}

\multirow{-2}{*}{CS16}& \textbf{UADA3D (ours) }    & \multicolumn{1}{c|}{\textbf{35.32}/ \textbf{45.53}} & \multicolumn{1}{c|}{{\color[HTML]{004000}    \textbf{15.21} / {\color[HTML]{004000}    \textbf{20.27}}}} & \multicolumn{1}{c|}{{\color[HTML]{004000}    \textbf{54.87} \%} / {\color[HTML]{004000}    \textbf{66.47} \%}}  & \multicolumn{1}{c|}{\underline{43.59} / \underline{52.93}} & \multicolumn{1}{c|}{{\color[HTML]{004000}    21.26}/\color[HTML]{004000}    14.85} & {{\color[HTML]{004000} 72.94\%} / \color[HTML]{004000} 69.04\%} \\ \cline{2-8} 
           
 & Oracle                    & \multicolumn{1}{c|}{47.82 / 55.76}          & \multicolumn{1}{c|}{-/-}                                      & \multicolumn{1}{c|}{-/-}           & \multicolumn{1}{c|}{51.48 / 59.58}          & \multicolumn{1}{c|}{-/-}                                      &  -/-             \\ \hline \hline \hline \hline

& Source Only               & \multicolumn{1}{c|}{7.28 / 12.93}          & \multicolumn{1}{c|}{-/-}    & -/-                                 & \multicolumn{1}{|c|}{5.38 / 25.97}          & \multicolumn{1}{c|}{-/-}                                      & -/- \\ \cline{2-8} 

& ST3D++~\cite{yang2022st3d++}  & \multicolumn{1}{|c|}{31.79 / 42.19}          & \multicolumn{1}{c|}{{\color[HTML]{004000}    24.50} / {\color[HTML]{004000}    29.26}}          &  \multicolumn{1}{c|}{{\color[HTML]{004000}    66.53 \%} / {\color[HTML]{004000}    74.69 \%}} & \multicolumn{1}{c|}{23.30 / 33.99}          & \multicolumn{1}{c|}{{\color[HTML]{004000}    17.92} / {\color[HTML]{004000}    8.01}}     & {\color[HTML]{004000}    41.67 \%} / {\color[HTML]{004000}    23.88 \%}       \\ \cline{2-8} 

\multirow{-2}{*}{CS64} & DTS~\cite{hu2023density}                        & \multicolumn{1}{c|}{\textbf{33.68} / \textbf{46.19}}               & {\color[HTML]{004000}    \textbf{26.40}} / {\color[HTML]{004000}    \textbf{33.26}}                          & \multicolumn{1}{|c}{{\color[HTML]{004000}    \textbf{71.69} \%} / {\color[HTML]{004000}    \textbf{84.90} \%}} & \multicolumn{1}{|c|}{16.87 / 32.13}               & \multicolumn{1}{c|}{{\color[HTML]{004000}    11.50} / {\color[HTML]{004000}    6.16}}          & {\color[HTML]{004000}    26.73 \%} / {\color[HTML]{004000}    18.34 \%} \\ \cline{2-8} 

\multirow{-2}{*}{\mydownarrow}& L.D.~\cite{wei2022lidar}                        & \multicolumn{1}{c}{20.70 / 32.10}               & \multicolumn{1}{|c|}{{\color[HTML]{004000}    13.41} / {\color[HTML]{004000}    19.17}}                        & \multicolumn{1}{c|}{{\color[HTML]{004000}    36.42 \%} / {\color[HTML]{004000}   48.94 \%}}  & \multicolumn{1}{|c|}{\underline{28.13} / \underline{38.90}}               & \multicolumn{1}{c|}{{\color[HTML]{004000}    22.75} / {\color[HTML]{004000}    12.93}}          & {\color[HTML]{004000}    52.89 \%} / {\color[HTML]{004000}    38.51 \%} \\ \cline{2-8}

\multirow{-2}{*}{R}& MS3D++~\cite{tsai2023ms3d++}                        & \multicolumn{1}{c|}{$\times$}               & \multicolumn{1}{c|}{$\times$}                         & \multicolumn{1}{c|}{$\times$}  & \multicolumn{1}{|c|}{14.44 / 37.88}               & \multicolumn{1}{c|}{{\color[HTML]{004000}    9.06} / {\color[HTML]{004000}    11.90}}          & {\color[HTML]{004000}    21.06 \%} / {\color[HTML]{004000}    35.47 \%} \\ \cline{2-8}

 & \textbf{UADA3D (ours)}    & \multicolumn{1}{c|}{\underline{33.00} / \underline{43.02}} & \multicolumn{1}{c|}{{\color[HTML]{004000} 25.72  } / {\color[HTML]{004000}   30.09 }}  &  \multicolumn{1}{c|}{{\color[HTML]{004000}   69.83 \%} / {\color[HTML]{004000}    76.81 \%}} & \multicolumn{1}{|c|}{\textbf{28.87} / \textbf{40.09}} & \multicolumn{1}{c|}{{\color[HTML]{004000}    \textbf{23.49}} / {\color[HTML]{004000}    \textbf{14.12}}} & {\color[HTML]{004000}    \textbf{54.62} \%} / {\color[HTML]{004000} \textbf{42.07} \%}  \\ \hline \hline


& Source Only               & \multicolumn{1}{c|}{$\times$}          & \multicolumn{1}{c|}{$\times$}    & $\times$                                 & \multicolumn{1}{|c|}{24.40 / 33.81}          & \multicolumn{1}{c|}{-/-}                                      & -/- \\ \cline{2-8} 

& ST3D++~\cite{yang2022st3d++}                    & \multicolumn{1}{c|}{$\times$}          & \multicolumn{1}{c|}{$\times$}     & $\times$       & \multicolumn{1}{|c|}{31.57 / 35.51}          & \multicolumn{1}{c|}{{\color[HTML]{004000}    7.17} / {\color[HTML]{004000}   1.70}}          & {\color[HTML]{004000} 16.19 \%} / {\color[HTML]{004000} 4.58 \%}  \\ \cline{2-8}  

\multirow{-2}{*}{K} & DTS~\cite{hu2023density}                        & \multicolumn{1}{c|}{$\times$}               & $\times$                     & \multicolumn{1}{|c}{$\times$}  & \multicolumn{1}{|c|}{23.24 / 23.56}               & \multicolumn{1}{c|}{{\color[HTML]{FE0000}    -1.16} / {\color[HTML]{FE0000}    -10.25}}          & {\color[HTML]{FE0000} -2.63 \%} / {\color[HTML]{FE0000} -27.69 \%} \\ \cline{2-8} 

\multirow{-2}{*}{\mydownarrow}& L.D.~\cite{wei2022lidar}                        & \multicolumn{1}{c}{$\times$}               & \multicolumn{1}{|c|}{$\times$}                        & $\times$  & \multicolumn{1}{|c|}{\underline{39.46} / \underline{41.47}}               & \multicolumn{1}{c|}{{\color[HTML]{004000}    15.06} / {\color[HTML]{004000}    7.66}}          & {\color[HTML]{004000} 33.98 \%} / {\color[HTML]{004000} 20.71 \%}  \\ \cline{2-8}

\multirow{-2}{*}{R}& MS3D++~\cite{tsai2023ms3d++}                        & \multicolumn{1}{c|}{$\times$}               & \multicolumn{1}{c|}{$\times$}                         & \multicolumn{1}{c|}{$\times$}  & \multicolumn{1}{|c|}{21.58 / 25.72}               & \multicolumn{1}{c|}{{\color[HTML]{FE0000}    -2.28} / {\color[HTML]{FE0000}    -8.09}}          & {\color[HTML]{FE0000} -6.36 \%} / {\color[HTML]{FE0000} -21.85 \%}  \\ \cline{2-8} 
& \textbf{UADA3D (ours) }    & \multicolumn{1}{c|}{$\times$} & \multicolumn{1}{c|}{$\times$}  & $\times$ & \multicolumn{1}{|c|}{\textbf{40.08} / \textbf{41.78}} & \multicolumn{1}{c|}{{\color[HTML]{004000}    \textbf{16.40}} / {\color[HTML]{004000}    \textbf{7.97}}} &  {\color[HTML]{004000} \textbf{37.01 \%}} / {\color[HTML]{004000} \textbf{21.54 \%}}  \\ \hline \hline


& Source Only               & \multicolumn{1}{c|}{$\times$}          & \multicolumn{1}{c|}{$\times$}    & $\times$                                 & \multicolumn{1}{|c|}{7.45 / 33.39}          & \multicolumn{1}{c|}{-/-}                                      & -/- \\ \cline{2-8} 

& ST3D++~\cite{yang2022st3d++}                    & \multicolumn{1}{c|}{$\times$}          & \multicolumn{1}{c|}{$\times$}     & $\times$       & \multicolumn{1}{|c|}{13.69 / 38.58}          & \multicolumn{1}{c|}{{\color[HTML]{004000}    6.24} / {\color[HTML]{004000}   5.19}}          & {\color[HTML]{004000} 15.24 \%} / {\color[HTML]{004000} 19.84 \%}  \\ \cline{2-8}  

\multirow{-2}{*}{N} & DTS~\cite{hu2023density}                        & \multicolumn{1}{c|}{$\times$}               & $\times$                     & \multicolumn{1}{|c}{$\times$}  & \multicolumn{1}{|c|}{11.25 / 34.75}               & \multicolumn{1}{c|}{{\color[HTML]{FE0000}    3.8} / {\color[HTML]{FE0000}  1.36}}          & {\color[HTML]{004000} 9.29 \%} / {\color[HTML]{004000} 5.20 \%} \\ \cline{2-8} 

\multirow{-2}{*}{\mydownarrow}& L.D.~\cite{wei2022lidar}                        & \multicolumn{1}{c}{$\times$}               & \multicolumn{1}{|c|}{$\times$}                        & $\times$  & \multicolumn{1}{|c|}{16.00 / 38.50}               & \multicolumn{1}{c|}{{\color[HTML]{004000}    8.55} / {\color[HTML]{004000}    5.11}}          & {\color[HTML]{004000} 20.89 \%} / {\color[HTML]{004000} 19.54 \%}  \\ \cline{2-8}

\multirow{-2}{*}{R}& MS3D++~\cite{tsai2023ms3d++}                        & \multicolumn{1}{c|}{$\times$}               & \multicolumn{1}{c|}{$\times$}                         & \multicolumn{1}{c|}{$\times$}  & \multicolumn{1}{|c|}{15.51 / 36.63}               & \multicolumn{1}{c|}{{\color[HTML]{004000}    8.06} / {\color[HTML]{004000}    3.24}}          & {\color[HTML]{004000} 19.70 \%} / {\color[HTML]{004000} 12.39 \%}  \\ \cline{2-8}

& \textbf{UADA3D (ours) }    & \multicolumn{1}{c|}{$\times$} & \multicolumn{1}{c|}{$\times$}  & $\times$ & \multicolumn{1}{|c|}{\textbf{16.77} / \textbf{38.73}} & \multicolumn{1}{c|}{{\color[HTML]{004000}    \textbf{9.32}} / {\color[HTML]{004000}    \textbf{5.34}}} &  {\color[HTML]{004000} \textbf{22.76 \%}} / {\color[HTML]{004000} \textbf{20.42 \%}}  \\ \hline \hline

 
 & Source Only               & \multicolumn{1}{c|}{8.33 / 15.93}          & \multicolumn{1}{c|}{-/-}                                    & \multicolumn{1}{c|}{-/-} &  \multicolumn{1}{c|}{3.78 / 6.00}          &      \multicolumn{1}{c|}{-/-} & -/-                             \\ \cline{2-8} 
 
 & ST3D++~\cite{yang2022st3d++} & \multicolumn{1}{c|}{\underline{21.57} / 31.04}          & \multicolumn{1}{c|}{{\color[HTML]{004000}    13.25} / {\color[HTML]{004000}    15.11}}          & \multicolumn{1}{c|}{{\color[HTML]{004000} 37.02 \%} / {\color[HTML]{004000} 41.77 \%}} & \multicolumn{1}{c|}{26.57 / 40.68}          & \multicolumn{1}{c|}{{\color[HTML]{004000}    22.80} / {\color[HTML]{004000}    34.69}}    & {\color[HTML]{004000} 51.09 \%} / {\color[HTML]{004000} 64.79 \%} \\ \cline{2-8}

 \multirow{-2}{*}{CS32} & DTS~\cite{hu2023density}                       & \multicolumn{1}{c|}{15.37 / 26.94}              & \multicolumn{1}{c|}{{\color[HTML]{004000}    7.05} / {\color[HTML]{004000}    11.01}}         & \multicolumn{1}{c|}{{\color[HTML]{004000} 19.69 \%} / {\color[HTML]{004000} 30.45 \%}}                            & \multicolumn{1}{c|}{18.68 / 29.65}              & \multicolumn{1}{c|}{{\color[HTML]{004000}    14.91} / {\color[HTML]{004000}    23.65}}             & {\color[HTML]{004000} 33.41 \%} / {\color[HTML]{004000} 44.18 \%}  \\ \cline{2-8} 
 
 \multirow{-2}{*}{\mydownarrow}& L.D.~\cite{wei2022lidar}                        & \multicolumn{1}{c|}{19.43 / \underline{31.38}}               & {{\color[HTML]{004000}    11.10} / {\color[HTML]{004000}    15.45}}                         & \multicolumn{1}{|c}{{\color[HTML]{004000} 31.02 \%} / {\color[HTML]{004000} 42.70 \%}}  & \multicolumn{1}{|c|}{\underline{26.89} / 36.85}               & \multicolumn{1}{c|}{{\color[HTML]{004000}    23.12} / {\color[HTML]{004000}    30.85}}          & {\color[HTML]{004000} 51.82 \%} / {\color[HTML]{004000} 57.62 \%} \\ \cline{2-8}
 
 \multirow{-2}{*}{R}& MS3D++~\cite{tsai2023ms3d++}                        & \multicolumn{1}{c|}{$\times$}               & \multicolumn{1}{c|}{$\times$}                        & \multicolumn{1}{c|}{$\times$}  & \multicolumn{1}{|c|}{4.45 / \underline{42.46}}               & \multicolumn{1}{c|}{{\color[HTML]{004000}    0.68} / {\color[HTML]{004000}    36.46}}          & {\color[HTML]{004000} 1.52 \%} / {\color[HTML]{004000} 68.09 \%} \\ \cline{2-8}
 
& \textbf{UADA3D (ours) }    & \multicolumn{1}{c|}{\textbf{22.15}/ \textbf{31.90}} & \multicolumn{1}{c|}{{\color[HTML]{004000} \textbf{13.82}} / {\color[HTML]{004000}    \textbf{15.97}}} & \multicolumn{1}{c|}{{\color[HTML]{004000} \textbf{38.62} \%} / {\color[HTML]{004000} \textbf{44.14} \%}} & \multicolumn{1}{c|}{\textbf{31.54}/ \textbf{42.82}} & \multicolumn{1}{c|}{{\color[HTML]{004000}    \textbf{27.76}} / {\color[HTML]{004000}    \textbf{36.83}}} & {\color[HTML]{004000} \textbf{62.23} \%} / {\color[HTML]{004000} \textbf{68.78} \%} \\ \hline  \hline


& Source Only               & \multicolumn{1}{c|}{10.28 / 15.26}          & \multicolumn{1}{c|}{-/-}                                   & \multicolumn{1}{c|}{-/-}   & \multicolumn{1}{c|}{15.07 / 33.97}          &    \multicolumn{1}{c|}{-/-}                                   & -/- \\ \cline{2-8} 

& ST3D++~\cite{yang2022st3d++}              & \multicolumn{1}{c|}{9.58 / 29.83}          & \multicolumn{1}{c|}{{\color[HTML]{FE0000}     -0.7} / {\color[HTML]{004000}    14.56}}         & \multicolumn{1}{c|}{{\color[HTML]{FE0000} -2.07 \%} / {\color[HTML]{004000} 39.53 \%}}       & \multicolumn{1}{c|}{33.22 / 42.62}          & \multicolumn{1}{c|}{{\color[HTML]{004000}    18.15} / {\color[HTML]{004000}    8.65}}     &{\color[HTML]{004000} 54.46 \%} / {\color[HTML]{004000} 33.84 \%}      \\ \cline{2-8} 

 \multirow{-2}{*}{CS16} & DTS~\cite{hu2023density}                       & \multicolumn{1}{c|}{\underline{11.95} / \underline{27.59}}              & \multicolumn{1}{c|}{{\color[HTML]{004000}    1.67} / {\color[HTML]{004000}    12.32}}         & \multicolumn{1}{c|}{{\color[HTML]{004000} 4.93 \%} / {\color[HTML]{004000} 33.45 \%}}                            & \multicolumn{1}{c|}{17.35 / 34.34}              & \multicolumn{1}{c|}{{\color[HTML]{004000}    2.28} / {\color[HTML]{004000}    0.37}}             & {\color[HTML]{004000} 6.84 \%} / {\color[HTML]{004000} 1.45 \%}  \\ \cline{2-8} 

\multirow{-2}{*}{\mydownarrow}& L.D.~\cite{wei2022lidar}                        & \multicolumn{1}{c|}{9.48 / 23.27}               & \multicolumn{1}{c|}{{\color[HTML]{FE0000}    -0.81} / {\color[HTML]{004000}    8.01}} & \multicolumn{1}{c|}{{\color[HTML]{FE0000} -2.38 \%} / {\color[HTML]{004000} 33.45 \%}} & \multicolumn{1}{|c|}{27.13 / 40.19}               & \multicolumn{1}{c|}{{\color[HTML]{004000}    12.06} / {\color[HTML]{004000}    6.22}}          & {\color[HTML]{004000} 36.19 \%} / {\color[HTML]{004000} 24.32 \%} \\ \cline{2-8}

\multirow{-2}{*}{R}& MS3D++~\cite{tsai2023ms3d++}                        & \multicolumn{1}{c|}{$\times$}               & \multicolumn{1}{c|}{$\times$}                         & \multicolumn{1}{c|}{$\times$}  & \multicolumn{1}{|c|}{2.16 / 32.30}               & \multicolumn{1}{c|}{{\color[HTML]{FE0000}    -12.91} / {\color[HTML]{FE0000}    -1.67}}           &  {\color[HTML]{FE0000} -38.75 \%} / {\color[HTML]{FE0000} -6.52 \%} \\ \cline{2-8}

 & \textbf{UADA3D (ours)}    & \multicolumn{1}{c|}{\textbf{19.71} / \textbf{35.69}} & \multicolumn{1}{c|}{{\color[HTML]{004000}    \textbf{19.71}} / {\color[HTML]{004000}    \textbf{20.42}}} & \multicolumn{1}{c|}{{\color[HTML]{004000} \textbf{27.87} \%} / {\color[HTML]{004000} \textbf{55.44} \%}} & \multicolumn{1}{c|}{\textbf{41.47} / \textbf{46.86}} & \multicolumn{1}{c|}{{\color[HTML]{004000}    \textbf{26.40}} / {\color[HTML]{004000}    \textbf{12.89}}} & {\color[HTML]{004000} \textbf{79.24} \%} / {\color[HTML]{004000} \textbf{50.43} \%} \\ \hline  \hline



 Robot & Oracle (R)                     & \multicolumn{1}{c|}{44.11 / 52.10}          & \multicolumn{1}{c|}{-/-}                                      & \multicolumn{1}{c|}{-/-}           & \multicolumn{1}{c|}{48.39 / 59.54}          & \multicolumn{1}{c|}{-/-}                                      & -/-

\end{tabular}
}

\end{table*}

\subsection{Evaluation metrics}
We report $mAP_{3D}$ and $mAP_{BEV}$, \textit{closed gap} and change in $mAP$. $mAP$ is the mean average precision over the three classes: Vehicle, Pedestrian, and Cyclist. \textit{Closed gap}~\cite{yang2019std} is calculated as $\frac{mAP_{model}-mAP_{source}}{mAP_{oracle}-mAP_{source}} \times 100\%$ and tells how close we are to a model trained on the target domain (oracle).

\section{Results}
\label{sec:results}

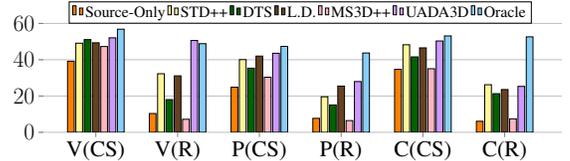
\begin{figure}[t!]
         \centering
\definecolor{byzantium}{rgb}{0.44, 0.16, 0.39}
\definecolor{arylideyellow}{rgb}{0.91, 0.84, 0.42}
\definecolor{paleaqua}{rgb}{0.69, 0.93, 0.93}
\definecolor{soft_orange}{rgb}{1.0, 0.7, 0.4}
\definecolor{sky_blue}{rgb}{0.53, 0.81, 0.92}
\definecolor{sage_green}{rgb}{0.74, 0.83, 0.6}
\definecolor{lavender}{rgb}{0.8, 0.6, 1.0}
\definecolor{soft_pink}{rgb}{1.0, 0.71, 0.76}
\definecolor{pale_yellow}{rgb}{1.0, 0.97, 0.6}
\definecolor{teal}{rgb}{0.53, 0.81, 0.98}
\definecolor{soft_coral}{rgb}{0.0, 0.39, 0.0}
\definecolor{dark_brown}{rgb}{0.4, 0.26, 0.13}

\begin{tikzpicture}

\begin{axis} [ybar = .05cm,
    bar width = 6pt,
    xmin = 0, 
    xmax = 5, 
    ymin=0,
    ymax=60,
    xshift=1.2cm,
    xtick pos=left,
    ytick pos=left,
    axis line style={draw=none},
    enlarge x limits = {abs=0.7},
    xtick = {0,1,2,3,4,5},
    xticklabels = {V(CS), V(R), P(CS), P(R), C(CS), C(R)},
    tick label style={font=\huge},
    legend style={font=\Large},
    label style={font=\hugehuge},
    y label style={at={(axis description cs:0,.5)}},
    legend style={scale=1,legend columns=-1, anchor=north, at={(0.5,1.2)}},
    height=5cm,width=18cm,
    ymajorgrids
]

\addplot[fill=orange] coordinates {(0,39.17) (2,24.85) (4, 34.72) 
(1,10.36) (3,7.7) (5, 6.08)}; 
\addplot[fill=pale_yellow] coordinates {(0,49.12) (2,40.07) (4, 48.25) 
(1,32.28) (3,19.58) (5,26.23)}; 
\addplot[fill=soft_coral] coordinates {(0,51.09) (2,35.34) (4, 41.57) 
(1,18.02) (3,15.08) (5, 21.29)};
\addplot[fill=dark_brown] coordinates {(0,49.33) (2,42.01) (4, 46.54) 
(1,31.12) (3,25.47) (5, 23.57)}; 
\addplot[fill=soft_pink] coordinates {(0,47.28) (2,30.36) (4, 35.04) 
(1,7.18) (3,6.45) (5, 7.42)}; 
\addplot[fill=lavender] coordinates {(0,52.07) (2,43.57) (4, 50.33) 
(1,50.60) (3,27.97) (5, 25.31)}; 
\addplot[fill=teal] coordinates {(0,56.83) (2,47.35) (4, 53.12) (1,48.85) (3,43.71) (5,52.61)}; 

\legend {Source-Only, STD++, DTS, L.D.,  MS3D++, UADA3D, Oracle};

\end{axis}

\end{tikzpicture}
         \vspace{-0.2cm}
         \caption{Per class $AP_{3D}$ in adaptation experiments Centerpoint. \textbf{V}ehicle, \textbf{P}edestrian, and \textbf{C}yclist. \textcolor{lavender}{\textbf{UADA3D}} is our method.}
         \label{fig:uda_class}
\end{figure}

In \cref{tab:adaptationMain} and \cref{tab:upsampling}, we compare our method with SOTA on a large number of UDA scenarios between source and target (S $\rightarrow$ T) data. In \cref{tab:adaptationMain} we focus on dense to sparse scenarios for the two models: Centerpoint~\cite{yin2021center} and IA-SSD~\cite{zhang2022not}. Our method, UADA3D, outperforms other SOTA approaches in most cases achieving much higher improvements, especially on the larger domain gaps (adapting models towards mobile R(obot) or W$ \rightarrow$ N). By analyzing \textit{Change} and \textit{Closed Gap} columns, we can observe that Centerpoint shows higher adaptability and improvement ($mAP_{3D}$ and $mAP_{BEV}$) than IA-SSD. That may come from the fact that, for IA-SSD, we have to fix a specific number of sampling points, which makes the model less flexible when adapting across different domains. Furthermore, even though other SOTA methods may outperform UADA3D on individual classes (only Cyclist in adaptation towards \textit{R}), they perform worse in others categories (\cref{fig:uda_class}). Thus, UADA3D generalizes better across different detectors and classes.


We observe that our method handles diverse domain shifts effectively when adapting between autonomous driving datasets (Waymo, KITTI, nuScenes), simulation data (LiDAR-CS), and robot data (R). As mentioned in \cref{sec:methods}, UADA3D does not need a pre-trained teacher model, as all the other approaches do. Instead, we can directly train the domain-adapted model, leveraging the GRL functionality which creates domain-invariant features. This allows our method to successfully train high-performing models on unlabeled target data, without depending on pseudo labels. We can observe in~\cref{tab:adaptationMain} that some of the methods perform even worse than the source-only approaches when tested on adaptation towards more challenging scenarios (e.g K $\rightarrow$ R), failing to generate accurate pseudo labels or distill teacher knowledge.Additionally, it is important to note that UADA3D never achieves lower performance than source-only models, on the adaptation task, regardless if the domain gap is big (e.g., K$\rightarrow$ R, W $\rightarrow$ N) or smaller (e.g., C64 $\rightarrow$ CS16). Notably, K$\rightarrow$R or CS16$\rightarrow$R adaptations appear particularly difficult, as many methods perform worse than source-only, suggesting that adapting to the robot's environment poses distinct challenges. However, despite this, the gap in those cases seems easier to close than the domain gap between W $\rightarrow$ N, where improvement and \textit{closed gap} were the lowest. 


In \cref{tab:upsampling}, we compare methods' performance on sparse to dense cases. UADA3D consistently outperforms other methods in all scenarios. All methods excel when adapting from sparse to dense data, indicating that target data is richer in information, compared to the source. This supports our hypothesis that adapting from dense to sparse is more challenging. Our method performs well on both: sparse to dense and dense to sparse, while some SOTA methods, like L.D.~\cite{wei2022lidar}, only allow adaptation from dense to sparse LiDAR, highlighting our method's generalizability.

\textbf{Additional Comparisons:} While we use Centerpoint and IA-SSD, our approach works with other detectors. In \cref{tab:upsampling}, we compare SECOND using weights provided by ST3D++, DTS, and MS3D++. The main limitation with using pre-trained models is that they are trained only on one category (car). Results highlight that UADA3D outperforms SOTA in these scenarios as well. Recently released CMDA~\cite{chang2024cmda} also uses an adversarial component, however, it is trained on multimodal data, while we only use LiDAR. We can compare our results in \cref{tab:upsampling} with CMDA results with SECOND reported on car: N-K (68.95/82.13) and W-N (24.64/42.81). Despite CMDA using additional modality, UADA3D performs on par with CMDA N-K and outperforms it on W-N.

    

\definecolor{LightCyan}{rgb}{0.849,1,0.969}

\begin{table*}[t]
\centering
\caption{Comparison of performance of different adaptation scenarios: CenterPoint on sparse to dense and various scenarios on SECOND~\cite{yan2018second}, previously used by SOTA authors. $mAP_{3D/BEV}$ reported over 3 classes: Vehicle, Pedestrian, and Cyclist on Centerpoint, and AP on Vehicle/Car class for SECOND. *Note: for SECOND we use weights provided by the authors.}
\label{tab:upsampling}

\resizebox{\textwidth}{!}{
\begin{threeparttable}
\begin{tabular}{c|cc|cc|cc|cc|cc}
                              & \multicolumn{2}{c|}{N \MVRightarrow K$^*$ (SECOND~\cite{yan2018second})}                                                               & \multicolumn{2}{c|}{W \MVRightarrow N$^*$(SECOND~\cite{yan2018second})} & \multicolumn{2}{c|}{W \MVRightarrow K$^*$(SECOND~\cite{yan2018second})}&\multicolumn{2}{c|}{N \MVRightarrow W (Centerpoint~\cite{yin2021center})} &\multicolumn{2}{c}{N \MVRightarrow K (Centerpoint~\cite{yin2021center})} \\ \cline{2-11}
                              & \multicolumn{1}{c|}{3D/BEV} & \multicolumn{1}{c|}{Closed Gap} & \multicolumn{1}{c|}{3D/BEV}                                          & Closed Gap & \multicolumn{1}{c|}{3D/BEV}                                          & Closed Gap & \multicolumn{1}{c|}{3D/BEV}                                         & Closed Gap & \multicolumn{1}{c|}{3D/BEV} & Closed Gap \\ \hline

 Source               & \multicolumn{1}{c|}{17.98/51.84}& \multicolumn{1}{c|}{-/-} & \multicolumn{1}{c|}{17.44/33.02}& \multicolumn{1}{c|}{-/-} & \multicolumn{1}{c|}{27.55/67.61}& \multicolumn{1}{c|}{-/-}                                    & \multicolumn{1}{c|}{19.61/37.76}        &\multicolumn{1}{c|}{-/-}  & \multicolumn{1}{c|}{17.18/40.18}                                     &   -/-     \\ \cline{2-11} 

 ST3D++                  & \multicolumn{1}{c|}{64.65/77.89}& \multicolumn{1}{c|}{84.14\% / 82.83\%}& \multicolumn{1}{c|}{22.03/36.71}& \multicolumn{1}{c|}{26.34\% / 19.56 \%}& \multicolumn{1}{c|}{\textbf{73.66}/83.37}& \multicolumn{1}{c|}{91.08\%/82.83\%}           & \multicolumn{1}{c|}{29.37/40.07} & \multicolumn{1}{c|}{{\color[HTML]{004000}    26.32\%}/{\color[HTML]{004000}   10.89\%}} & \multicolumn{1}{c|}{\underline{20.37}/44.02} & \multicolumn{1}{c}{{\color[HTML]{004000}11.66\%/36.99\%}}          \\ \cline{2-11} 

 DTS                       & \multicolumn{1}{c|}{\underline{64.97}/\underline{80.23}}& \multicolumn{1}{c|}{84.71 \% / 90.27\%}& \multicolumn{1}{c|}{22.83/40.62}& \multicolumn{1}{c|}{30.92\% / 40.29 \%}& \multicolumn{1}{c|}{71.18/\underline{83.91}}& \multicolumn{1}{c|}{91.08\%/82.83\%} & \multicolumn{1}{c|}{\underline{32.33}/\underline{41.10}}& \multicolumn{1}{c|}{{\color[HTML]{004000}34.31\%/15.77\%}}& \multicolumn{1}{c|}{{18.90/\underline{46.20}}}& \multicolumn{1}{c}{{\color[HTML]{004000}6.30\%/57.95\%}}           \\ \cline{2-11}

 MS3D++                       & \multicolumn{1}{c|}{X}& \multicolumn{1}{c|}{X}& \multicolumn{1}{c|}{23.15/\underline{44.01}}& \multicolumn{1}{c|}{32.75\% / 58.28 \%}& \multicolumn{1}{c|}{X}& \multicolumn{1}{c|}{X} & \multicolumn{1}{c|}{21.82/39.92} & \multicolumn{1}{c|}{{\color[HTML]{004000} 5.97\%/10.19\%}}& \multicolumn{1}{c|}{18.00/40.61}& \multicolumn{1}{c}{{\color[HTML]{004000}2.98\%/4.41\%}} \\ \cline{2-11}

 \textbf{UADA3D}     & \multicolumn{1}{c|}{\textbf{65.91}/\textbf{82.44}}& \multicolumn{1}{c|}{86.41\% / 97.30\%}& \multicolumn{1}{c|}{\textbf{24.84}/\textbf{45.51}}& \multicolumn{1}{c|}{42.44\% / 66.21 \%}& \multicolumn{1}{c|}{\underline{73.41}/\textbf{86.53}}& \multicolumn{1}{c|}{95.74\%/96.14\%} & \multicolumn{1}{c|}{\textbf{35.68}/\textbf{42.78}} & \multicolumn{1}{c|}{{\color[HTML]{004000}\textbf{43.33}\%/\textbf{23.72}\%}} & \multicolumn{1}{c|}{\textbf{21.97}/\textbf{48.52}} & \multicolumn{1}{c}{{\color[HTML]{004000}\textbf{17.51}\%/\textbf{80.35}\%}} \\ \cline{2-11} 

 Oracle                    &  \multicolumn{1}{c|}{73.45/83.29}& \multicolumn{1}{c|}{-/-}& \multicolumn{1}{c|}{34.87/51.88}& \multicolumn{1}{c|}{-/-}& \multicolumn{1}{c|}{75.45/87.29}& \multicolumn{1}{c|}{-/-} 
 & \multicolumn{1}{c|}{56.69/58.96} & \multicolumn{1}{c|}{-/-}  & \multicolumn{1}{c|}{44.51/50.56} & \multicolumn{1}{c}{-/-}            \\

\end{tabular}
  \end{threeparttable}
}
\end{table*}



\subsection{Ablation Studies}
\label{sec:ablstudies}
First, we ask the question of which probability distribution alignment is the most beneficial for UADA3D. Next, we investigate the impact of discriminator design and analyze GRL parameters. Finally, we integrate other self-learning components to enhance our method further. 





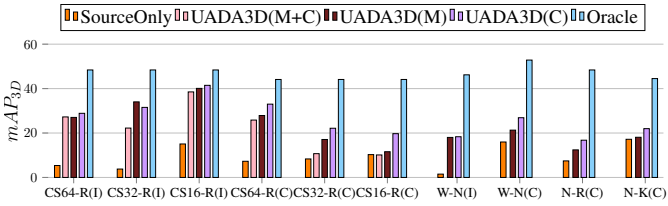
\begin{figure}[h!]
         \centering
         \definecolor{persianplum}{rgb}{0.44, 0.11, 0.11}
\definecolor{byzantium}{rgb}{0.44, 0.16, 0.39}
\definecolor{operamauve}{rgb}{0.72, 0.52, 0.65}
\definecolor{paleaqua}{rgb}{0.69, 0.93, 0.93}
\definecolor{teal}{rgb}{0.53, 0.81, 0.98}
\definecolor{lavender}{rgb}{0.8, 0.6, 1.0}
\definecolor{soft_pink}{rgb}{1.0, 0.71, 0.76}

\begin{tikzpicture}
\begin{axis} [
    name=ax1,
    bar width = 4pt,
    xmin = -0.1, 
    xmax = 9, 
    ybar = 2pt,
    ymin=0,
    ymax=60,
    xtick pos=left,
    ytick pos=left,
    axis line style={draw=none},
    enlarge x limits = {abs=0.4},
    xtick = data,
    xticklabels = {CS64-R(I), CS32-R(I), CS16-R(I), CS64-R(C), CS32-R(C), CS16-R(C), W-N(I), W-N(C), N-R(C), N-K(C)},
    ylabel={$mAP_{3D}$},
    tick label style={font=\normalsize},
    legend style={font=\Large},
    label style={font=\Large},
    y label style={at={(axis description cs:0.05,0.5)},anchor=south},
    legend style={scale=1,legend columns=-1, anchor=north, at={(0.5,1.3)}},
    nodes near coords style={font=\normalsize},
    height=5cm,width=17.5cm,
    ymajorgrids
]

\addlegendentry{SourceOnly}
\addplot[fill=orange] coordinates {(0,5.38) (1,3.78) (2,15.07) (3,7.28) (4,8.33) (5, 10.28) (6.11, 1.5) (7.11, 15.96) (8.11, 7.45) (9.11, 17.18)}; 

\addlegendentry{UADA3D\textsubscript{$\mathcal{L}mc$}}
\addplot[fill=soft_pink] coordinates {(0,27.25) (1, 22.23) (2,38.51) (3,25.83) (4, 10.71) (5,10.13) }; 

\addlegendentry{UADA3D\textsubscript{$\mathcal{L}m$}}
\addplot[fill=persianplum] coordinates {(0,27.03) (1, 34.02) (2,40.1) (3,27.88) (4, 17.09) (5,11.58) (6, 18.02) (7, 21.30) (8, 12.41) (9, 18.09)}; 

\addlegendentry{UADA3D}
\addplot[fill=lavender] coordinates {(0,28.87) (1, 31.54) (2,41.47) (3,33.00) (4, 22.15) (5,19.71) (6, 18.33) (7, 26.89) (8, 16.77) (9, 21.97)}; 

\addlegendentry{Oracle}
\addplot[fill=teal]  coordinates {(0, 48.39) (1, 48.39) (2, 48.39) (3, 44.11) (4, 44.11) (5, 44.11) (6, 46.20) (7, 52.84) (8, 48.39) (9, 44.51)}; 

\legend {SourceOnly, UADA3D(M+C), UADA3D(M), UADA3D(C),  Oracle};
 
\end{axis}

  
\end{tikzpicture}
         \vspace{-0.1cm}
         \caption{Comparison UADA3D performance with different probability distribution alignments on IA-SSD (I) and Centerpoint (C).}
         \label{fig:UADA_IASSD_Ab}
\end{figure}

\textbf{Probability Distribution Alignments:} We explore the effectiveness of other probability distribution alignment strategies, shown in~\cref{fig:UADA_IASSD_Ab}. In UADA3D with marginal distribution alignment, UADA3D\textsubscript{$\mathcal{L}m$}, the discriminator gradient is backpropagated only to the feature extractor with loss \(\mathcal{L}_m\), where $\mathcal{L}_{m} = \frac{1}{N} \sum_{n=1}^N  d \cdot \log (g_{\theta_m}(X_n) ) + (1 - d) \cdot \log ( 1 - g_{\theta_m}(X_n))$, where $d$ is 0 for source and 1 for target domains, and $g_{\theta_M}$ denotes the marginal discriminator network. In UADA3D (i.e., UADA3D\textsubscript{$\mathcal{L}c$}) the discriminator gradient is backpropagated to the detection head through the whole model. UADA3D\textsubscript{$\mathcal{L}mc$} combines marginal and conditional alignment with
\(\mathcal{L}_{mc} =  (\mathcal{L}_{m}+\mathcal{L}_{C})\), where $\mathcal{L}_{C}$ is given in \cref{eq:loss_cond}.
UADA3D\textsubscript{$\mathcal{L}m$} uses feature maps directly, without features masking, to predict the domain and calculate the loss, while UADA3D employs masked features with class labels and bounding boxes.
UADA3D with conditional probability distribution alignment (and feature masking) consistently yields higher-quality outcomes, however UADA3D\textsubscript{$\mathcal{L}m$} delivers comparable results in cross-sensor adaptation for self-driving cars, due to significant differences in the marginal probability distribution.

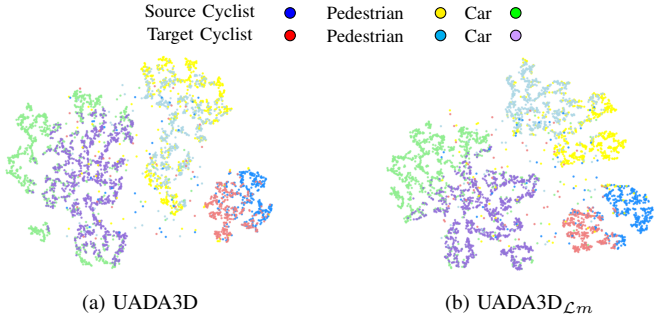
\begin{figure}[h]
      \begin{tikzpicture}
        \scriptsize
        \node[draw, fill=blue, circle, inner sep=0pt, minimum size=4pt] at (0, 0) {};
        \node at (-1.2, 0) {Source Cyclist};
        
        \node[draw, fill=yellow, circle, inner sep=0pt, minimum size=4pt] at (2, 0) {};
        \node at (1, 0) {Pedestrian};
        
        \node[draw, fill=green, circle, inner sep=0pt, minimum size=4pt] at (3, 0) {};
        \node at (2.5, 0) {Car};
        
        \node[draw, fill=red, circle, inner sep=0pt, minimum size=4pt] at (0, -0.3) {};
        \node at (-1.2, -0.3) {Target Cyclist};
        
        \node[draw, fill=cyan, circle, inner sep=0pt, minimum size=4pt] at (2, -0.3) {};
        \node at (1, -0.3) {Pedestrian};
        
        \node[draw, fill=lavender, circle, inner sep=0pt, minimum size=4pt] at (3, -0.3) {};
        \node at (2.5, -0.3) {Car};
    \end{tikzpicture}
    \centering
    \begin{subfigure}[b]{0.43\columnwidth}
        \centering
        \includegraphics[width=\textwidth]{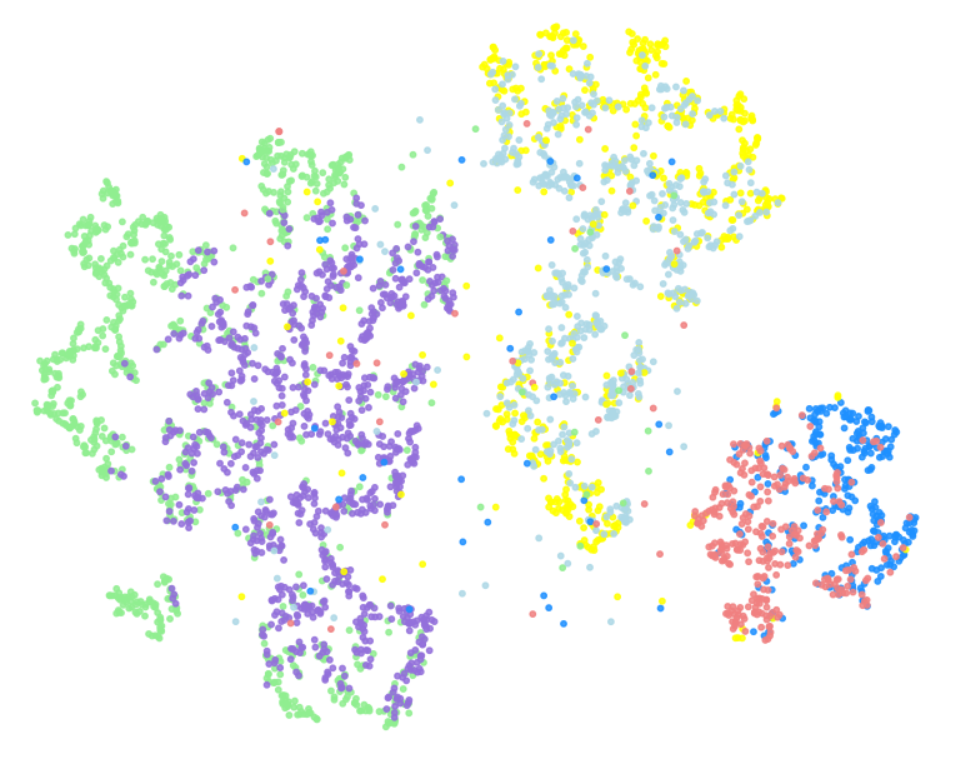}
        \caption{UADA3D }
        \label{fig:tsneUADA}
    \end{subfigure}
    \hfill
    \begin{subfigure}[b]{0.43\columnwidth}
        \centering
        \includegraphics[width=\textwidth]{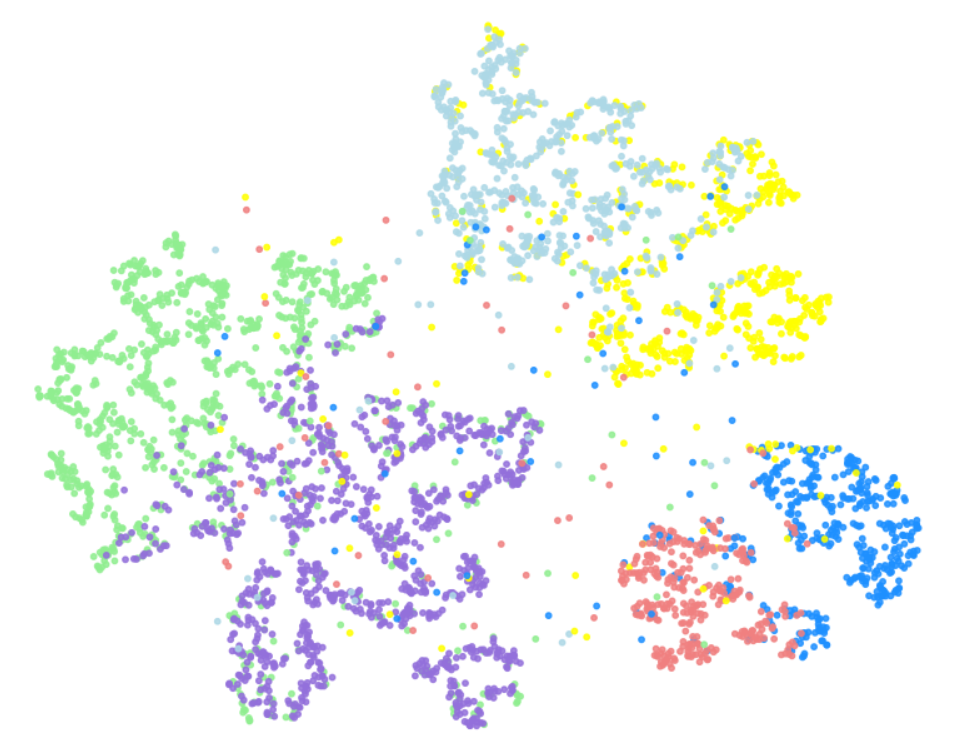}
        \caption{UADA3D\textsubscript{$\mathcal{L}m$}}
        \label{fig:tsneUADA_M}
    \end{subfigure}
    

    \caption{T-SNE analysis for N $\rightarrow$ Robot with Centerpoint.}
    \label{fig:tsne}
\end{figure}

We also want to bring up the importance of feature masking and per-class discrimination in UADA3D. Using T-SNE analysis~\cite{van2008visualizing}, we can observe in~\cref{fig:tsneUADA} that class wise discrimination with feature masking helps us achieve class-wise domain invariant features. If we compare this to features obtained with UADA3D\textsubscript{$\mathcal{L}m$} in \cref{fig:tsneUADA_M}, which uses the global features to predict the domain without feature masking we can observe that the class wise features are not well aligned. Consequently, we selected conditional alignment with feature masking as our preferred method.

    


\textbf{Discriminator designs:} In \cref{tab:multi-ablation} we examine different discriminator designs: (a) single domain discriminator with the input of $(x_n,\hat{b}_n)$ (features masked by predicted bounding boxes), without label information $\hat{y}$, (b) a single domain discriminator with the input of $(x_n,\hat{b}_n,\hat{y})$, where the discriminator output is multiplied by the maximum predicted class confidence $\hat{y}_{max,n} \odot g_{\theta_D}(x_n,\hat{b}_n)$, and (c) the default UADA3D setting with input of $(x_n,\hat{b}_n,\hat{y})$ and three class-wise domain discriminators, where the output of each discriminator is multiplied by the corresponding class confidence.  
(a) and (b) obtain higher adaptation scores in some cases, especially on Centerpoint CS64$\rightarrow$R, but worse than (c) on IA-SSD. Comparing (a) and (b), we can see that the multiplication with $\hat{y}_{max}$ (b) has a substantial impact. Centerpoint obtains better performance using one discriminator than IA-SSD, which is likely due to the easier transferability of BEV features compared to point features. $AP_{BEV}$ scores of IA-SSD also improve more than the $AP_{3D}$ for cases (a) and (b). Due to higher model complexity, IA-SSD appears to gain the most advantage from the use of multiple class-wise conditional discriminators (c). While IA-SSD performs the best with (c) and Centerpoint with (b), to be consistent, we chose option (c) as a default option for our method.


\begin{table}[h]
\fontsize{65}{88}\selectfont 
\caption{Contribution of class-confidence and multiple discriminators ($AP_{3D}/AP_{BEV}$). $N_D$ is a number of class-wise discriminators, and $y$ denotes if we use class labels.}
\vspace{-0.1cm}
\label{tab:multi-ablation}
\centering
\resizebox{\columnwidth}{!}{
\begin{tabular}{@{}l|c|c|c|c|c|c|c|c|c|c|c|c@{}}
                             & & & & \multicolumn{3}{c|}{CS64 \MVRightarrow 
 CS16} & \multicolumn{3}{c}{CS64 \MVRightarrow robot} \\ \cline{5-10}
                              \multicolumn{1}{c|}{}        & M   & \textit{N\textsubscript{D}} & \textit{y}    & \multicolumn{1}{c|}{\textit{V\textsubscript{3D/BEV}}} & \textit{P\textsubscript{3D/BEV}} & \textit{C\textsubscript{3D/BEV}} & \multicolumn{1}{c|}{\textit{V\textsubscript{3D/BEV}}} & \textit{P\textsubscript{3D/BEV}} & \multicolumn{1}{c}{\textit{C\textsubscript{3D/BEV}}} \\ \hline
                              \multirow{2}{*}{IA}      & S &                   &              & 29.97 / 38.10   & 11.02 / 14.62  & 19.35 / 23.05  & 1.17 / 5.77   & 6.42 / 18.10  &  \multicolumn{1}{c}{14.26 / 14.92} \\
                                                           & (a)       & 1                 &              & 42.55 / 61.42  & 17.36 / 32.85  & 35.29 / 39.14   & 52.20 / 59.60  & 8.28 / 20.65  & \multicolumn{1}{c}{15.47 / 23.37}  \\
                                                        \multirow{2}{*}{\cite{zhang2022not}}     & (b)       & 1                 & $\checkmark$ & 40.92 / 61.61  & 17.75 / 32.51  & 34.99 / 39.81  & \textbf{59.85} / \textbf{67.66}   & 11.32 / 25.70  & \multicolumn{1}{c}{14.01 / 16.51}  \\
                                                        & (c)       & 3                 & $\checkmark$ & \textbf{42.96} / \textbf{62.08 } & \textbf{25.08} / \textbf{34.09}  & \textbf{37.91} / \textbf{40.42}  & 53.29 / 63.93  & \textbf{14.20} / \textbf{28.58}  & \multicolumn{1}{c}{\textbf{31.51} / \textbf{36.55}}  \\ \hline 
                         \multirow{2}{*}{Cent} & S &                   &              & 32.30 / 48.08  & 11.63 / 20.09  & 23.06 / 46.07  & 1.11 / 48.16  & 3.03 / 14.06  & \multicolumn{1}{c}{11.99 / 15.70}  \\
                                                        & (a)       & 1                 &              & \textbf{47.95} / 65.84  & 39.21 / 49.00  & 45.91 / 47.38  & 46.94 / 54.89  & 25.04 / 44.12  & \multicolumn{1}{c}{39.43 / 44.57}  \\
                                                \multirow{2}{*}{\cite{yin2021center}}            & (b)       & 1                 & $\checkmark$ & 47.62 / 65.43  & \textbf{40.28} / 49.42  & \textbf{46.08} / 47.41  &\textbf{50.81} /\textbf{58.72}  & \textbf{27.68} / \textbf{45.39}  & \multicolumn{1}{c}{\textbf{47.88} / \textbf{51.56}}  \\
                                                           & (c)       & 3                 & $\checkmark$ & 46.99 / \textbf{66.01}  & 38.22 / \textbf{49.47}  & 45.56 / \textbf{48.30}   & 44.53 / 53.11  & 20.65 / 33.50  & \multicolumn{1}{c}{21.43 / 33.67} 

\end{tabular}
}
\end{table}


\textbf{Gradient reversal coefficient:}
We tested two different strategies for \textit{GRL}-coefficient $\lambda$ on UADA3D and UADA3D\textsubscript{$\mathcal{L}m$}. Firstly, a constant $\lambda=0.1$ was tested following the setting used for most adversarial UDA strategies in 2D object detection~\cite{chen2018domain,saito2019strong}. Secondly, we follow other approaches~\cite{ganin2015unsupervised,li2023domain} where $\lambda$ was increased over the training according to: $\lambda = \alpha (\frac{2}{1 + exp(-\gamma p)} - 1)$, where $\alpha \in [0,1]$ is a scaling factor that determines the final $\lambda$, $\gamma=10$ and $p$ is the training progress from start $0$ to finish $1$. $\alpha$-values of $1, 0.5, 0.2, 0.1$ were tested. In this way, the gradient reversal is low for the first few iterations and goes towards $\alpha$. Given that we do not use a pre-trained model, this approach enables the method to concentrate on the object detection task at the beginning of training, as the discriminator loss will initially be significantly smaller. This performance stems from the characteristics of each category. Pedestrians show more variability in posture and occlusion, while cars and cyclists are more consistent. We hypothesize that dynamic adaptation improves generalization for pedestrians, whereas for cars and cyclists, the impact of $\lambda$ is less significant. In \Cref{tab:ablation_grl}, we can observe that smaller $\lambda$ values result in better adaptation performance. The best score for vehicles and cyclists was achieved with the constant $\lambda=0.1$ while the dynamic $\lambda$ achieved the best scores for pedestrians at $\alpha=0.1$ respectively. Both $\lambda=0.1$ and $\alpha=0.1$ yielded good average scores across all three classes with minimal differences. We tested constant $\lambda \in [0.01,0.5]$ and observed a slow decrease in performance towards source-only values for smaller values and low performance for higher values thus we opted for constant $\lambda=0.1$ in our experiments.

\begin{table}[t]
    \centering
    \caption{GRL coefficients tests of IA-SSD, for \textit{CS64} $\rightarrow$ \textit{CS16}.}
    \vspace{-0.2cm}
    \resizebox{0.68\columnwidth}{!}{%
    \label{tab:ablation_grl}
    \begin{tabular}{c|ccc}
 $\lambda$                               & \multicolumn{3}{c}{UADA3D/UADA3D\textsubscript{$\mathcal{L}m$}}  \\ \hline
                                                                     & $AP_{3D, V}$ & $AP_{3D, P}$ & $AP_{3D, C}$ \\ \hline
                   Source                                       & 29.97        & 11.02        & 19.35        \\ \hline
                   0.05 & 38.13/37.04 & 23.41/22.13 & 29.29/26.85 \\
                     0.1                                     & 42.96/\textbf{48.48} & 25.48/25.08  & \textbf{38.12}/\textbf{37.91}  \\
                                        0.2 & \textbf{43.24}/40.34 & 24.86/24.91 & 34.24/32.96 \\ \hline

                            $\alpha=0.1$ & 42.76/41.06  & \textbf{26.43}/\textbf{26.80}  & 33.77/31.83  \\
                             $\alpha=0.2$ & 40.14/38.11  & 24.55/24.7   & 34.22/30.01  \\
                             $\alpha=0.5$ & 40.24/37.55  & 23.58/21.42  & 33.16/27.43  \\
                             $\alpha=1$      & 41.65/40.09  & 23.68/22.58  & 31.72/29.39  \\ \hline
                    Oracle                      & 58.62        & 35.57        & 49.27       
\end{tabular}}

\end{table}

\section{Conclusions and Limitations}
We introduce UADA3D, a novel approach tailored for challenging UDA scenarios, specifically addressing sparser LiDAR data and mobile robots. UADA3D effectively navigates diverse environments, ensuring precise object detection and achieving SOTA performance across various domain adaptation scenarios. This way, the large number of existing data sets in the field of autonomous driving can also be leveraged for mobile robotics. However, domain gaps persist, notably in scenarios such as W $\rightarrow$ N or K $\rightarrow$ R, where improvements are constrained. Future work could explore adapting between multimodal scenarios, such as camera and LiDAR, radar and LiDAR, or adding more self-training components.



{
    \footnotesize
    \bibliographystyle{ieeetr}
    \bibliography{ref}
}

\appendix

\section{Implementation Details}

\subsection{Detailed implementation of UADA3D}


The conditional module has the task of reducing the discrepancy between the conditional label distribution $P(Y_s|X_s)$ of the source and $P(Y_t|X_t)$ of the target. The label space $Y_i$ consists of class labels $y \in \mathbb{R}^{N \times K}$ and 3D bounding boxes $b_i \in \mathbb{R}^{7}$. The feature space $X$ consists of point features $F \in \mathbb{R}^{N\times C}$ (IA-SSD) or the $2$D BEV pseudo-image $I\in \mathbb{R}^{w \times h \times C}$ (Centerpoint). The domain discriminator $g_{\theta_D}$ in Centerpoint has $2$D convolutional layers of $264,256,128,1$ while IA-SSD uses an MLP with dimensions $519,512,256,128,1$. LeakyReLU is used for the activation functions with a sigmoid layer for the final domain prediction. 
A kernel size of $3$ was chosen for Centerpoint, based on experiments shown in  \ref{sub:specific_settings}. 
Note, that we do class-wise domain prediction, thus we have $K$ discriminators corresponding to the number of classes (in our case $K=3$, but it can be easily modified). 

subsection{Detailed implementation of UADA3D\textsubscript{$\mathcal{L}m$}}


The primary role $UADA3D_{\mathcal{L}m}$ with the marginal feature discriminator is to minimize the discrepancy between the marginal feature distributions of the source, denoted as \( P(X_s) \), and the target, represented by \( P(X_t) \). Here, \( X_s \) and \( X_t \) symbolize the features extracted by the detection backbone from the two distinct domains. This approach ensures the extraction of domain invariant features. The loss function of $UADA3D_{\mathcal{L}m}$ marginal alignment module is defined through Binary Cross Entropy.

The output of the point-based detection backbone in IA-SSD~\cite{zhang2022not} is given by $N$ point features with feature dimension $C$ and corresponding encodings. Point-wise discriminators can be utilized to identify the distribution these points are drawn from. The input to the proposed marginal discriminator $g_{\theta_D}$ is given by point-wise center features obtained through set abstraction and downsampling layers. The discriminator is made up of $5$ fully connected layers (512,256,128,64,32,1) that reduce the feature dimension from $C$ to $1$. LeakyReLU is used in the activation layers and a final sigmoid layer is used for domain prediction. 


The backbone in Centerpoint~\cite{yin2021center} uses sparse convolutions to extract voxel-based features that are flattened into 2D BEV-features. Therefore, the input to the view-based marginal discriminator is given by a pseudo image of feature dimension $C$ with spatial dimensions $w$ and $h$ that define the 2D BEV-grid. Since 2D convolutions are more computationally demanding over the MLP on the heavily downsampled point cloud in IA-SSD, the 2D marginal discriminator uses a $3$-layered CNN that reduces the feature dimension from $C$ to $1$ (256,256,128,1), using a kernel size of $3$ and a stride of $1$. Same as in the point-wise case, the loss function of $UADA3D_{\mathcal{L}m}$ is defined through Binary Cross Entropy.

\subsection{Memory and Runtime}

In this section, we tried to include all the details regarding training, model sizes, and inference time. First, we want to mention that our choice of models was driven by two main factors: how well-performing, adaptable, and accessible (open-source) the methods are (thus, we chose CenterPoint), but also whether the method would be a feasible choice for our mobile robot. Unfortunately, during initial tests, it became clear that while CenterPoint seems to be a better choice for autonomous driving applications, it was not the best choice for mobile robots. After further research, we added IA-SSD and found that, while its adaptability is not as good as CenterPoint’s, it achieves quite good performance and can run the perception models onboard the mobile robot (at inference time). We benchmarked our models on Laura (mobile robot) equipped with Jetson Xavier. Due to limited memory, we had to apply mixed precision and convert the models' weights to FP16.

We must mention that while it would be possible to run the domain adaptation and experiments entirely on the mobile platform, it would be infeasible. Experiments would take significantly more time due to the GPU capacity (sufficient for inference but much slower for training). Moreover, the bottleneck in this training would be data access. While on a cluster or local PC we have high-speed data storage, our mobile robot has only 256 GB of storage, and reading data from an external drive would be much slower.

\subsection{Model Size:}

\paragraph{CenterPoint} The parameter count of ~13 million. Model weights around 48 MB. Note, that we use lightweight configuration On an NVIDIA V100 GPU (32 GB), CenterPoint can process around 10-12 FPS (frames per second). With newer GPUs like the A100 or more aggressive optimizations (e.g., TensorRT or FP16 precision), you can get slightly better performance (~15-17 FPS). 

On Laura we were able to process 3-4 FPS. This is due to the computational complexity CenterPoint of 3D object detection models. Using TensorRT for inference and precision reduction (e.g., FP16 or INT8) we could push the throughput to 5-6 FPS.

\paragraph{IA-SSD} IA-SSD has a much smaller parameter count (~7 million) compared to heavier models, that yields an estimate of around 31 MB. IA-SSD is designed for efficiency. On an NVIDIA V100 (32 GB), IA-SSD can achieve 20-25 FPS. On more modern hardware (e.g., A100), we can go up to 30-35 FPS with optimizations. On Laura's Jetson Xavier we achieve 6-8 FPS and with TensorRT and FP16 or INT8 precision, we could reach 10-12 FPS.

We specifically used IA-SSD for mobile robot adaptation due to its balance between computational efficiency and adaptability. While IA-SSD may not perform as well as CenterPoint in autonomous driving scenarios, it is well-suited for onboard deployment on our mobile robot, allowing us to run the model efficiently at inference time.

\noindent
\paragraph{Training Details:} 
\paragraph{GPU Hours Calculation:} For CS-LiDAR, Kitti and Laura data 
\begin{itemize}
  \item \textbf{UADA3D CenterPoint:} 40 epochs, each model taking 10 hours.
  \item \textbf{UADA3D IA-SSD:} 80 epochs, each model taking 12 hours.
  \item \textbf{Oracle/Source-only Models:} CenterPoint (3.5 hours), 40 epochs and IA-SSD (4 hours), 80 epochs.
  \item \textbf{Comparison Methods:} For CenterPoint, approximately 8-10 hours; for IA-SSD, about 10-12 hours, 80 epochs for each model.
\end{itemize}

\noindent

For the Waymo and nuScenes datasets, which are substantially larger than other datasets, we estimated needing approximately 22 times more GPU hours to complete training on Waymo and 4 times more for nuScenes. To train multiple scenarios efficiently, we used a GPU cluster node with 8 A100 (40GB) GPUs. This allowed us to achieve comparable training times for nuScenes and approximately 4-5 times longer training times for Waymo.
Note that our training takes the same amount of time as the other comparison methods. However, we do not require the additional 3-4 hours of training for a teacher model as the other methods do. Additionally, in each epoch, we process more data than the comparison methods since we loop over both source and target, whereas the other methods only loop over the target (they only use the source during teacher training). This demonstrates that our algorithm is faster, as it achieves a similar runtime on a larger amount of data.

\section{Detectors' specific settings and ablation} \label{sub:specific_settings}
We ran different hyper-parameter studies of the two detectors in order to train the best source-only and oracle models for the methods we compared against. 

\subsection{IA-SSD sampling settings}
Three different sampling settings were tested for supervised training of the IA-SSD detector to find sampling settings appropriate for the data densities provided by 64, 32 and 16 layer LiDARs. Firstly, the sampling settings proposed by the authors~\cite{zhang2022not} for the \textit{KITTI} dataset where the points are progressively downsampled from $4096$ to $256$ points. Secondly, an intermediate with progressive downsampling from $8192$ to $512$ points. 
Finally, the authors' settings proposed for \textit{Waymo}~\cite{sun2020waymo} where downsampling is performed progressively from $16384$ to $1024$ points. The three tested settings with intermediate sampling layers are seen in \Cref{tbl:SA-layers}.

\begin{table}[h]
    \centering
    \caption{SA sampling settings for IA-SSD~\cite{zhang2022not}.}
    \resizebox{\columnwidth}{!}{
    \begin{tabular}{c|cccc}
        Sampling & \multicolumn{4}{c}{Layers}  \\
         Setting & \textit{D-FPS} & \textit{D-FPS} & \textit{ctr-aware} & \textit{ctr-aware} \\
         
        \hline
        1 & 4096 & 1024  & 512  & 256  \\
        2 & 8192 & 2048  & 1024 & 512  \\
        3 & 16384 & 4096 & 2048 & 1024 \\
    \end{tabular}}
    \label{tbl:SA-layers}
\end{table}

The number of points in the four sampling layers of IA-SSD~\cite{zhang2022not} was set according to experiments (see \Cref{tbl:params-sampling}). We selected the intermediate setting of $8192$, $2048$, $1024$ and $512$ for all runs including IA-SSD. This was chosen as a trade-off between detection accuracy and speed. The input point cloud was downsampled using random sampling to $65536$ in \textit{CS-64} following the author's implementation on Waymo~\cite{sun2020waymo} dataset that features a similar point cloud density, \textit{CS-32} was downsampled to $32768$, while $16384$ was chosen for \textit{CS-16} and \textit{Robot}.

\begin{table}[h]
\centering
\caption{Supervised results on the \textit{LiDAR-CS}~\cite{fang2023lidar} dataset using IA-SSD~\cite{zhang2022not}. with different sampling settings.}
\resizebox{\columnwidth}{!}{%
\begin{tabular}{ccc|cccc}
    Sensor   &  Sampling   & $n_{input}$ & $AP_{3D, V}$ & $AP_{3D, P}$ & $AP_{3D, C}$ & Speed \\
       & Setting  & & & &  & [it/s]\\

    \hline

          \textit{CS-64}  &  1  & 65536 &     76.51         &         17.35       &   47.71    &     20.2    \\
           &   2  & 65536 &       87.81       &       50.41       &    78.55   &    3.9    \\
          &  3  & 65536 &     \textbf{90.51}         &    \textbf{ 65.97}           &    \textbf{84.67}   &      1.9   \\
          

       \hline
                 \textit{CS-32}  &  1  & 32768 & 55.06             & 18.99               & 25.34    & 20.2         \\
           &   2  & 32768 &  63.41           & \textbf{46.50}             & 53.31       &    7.8    \\
          &  3  & 32768 &    \textbf{65.11}           &  42.89             & \textbf{54.03} &    3.6    \\
          

       \hline
\textit{CS-16}  &  1   &  16384 &    50.39        &      17.41         &       34.00    & 20.2   \\ 
            &  2  &  16384 &      \textbf{53.54}      &       \textbf{47.78}        &      \textbf{53.11}      & 13.9   \\ 
            &  3  &  16384 &     56.76      &       43.29        &       52.88     & 10.6    \\ 
     
\end{tabular}
}
\label{tbl:params-sampling}
\end{table}


\subsection{Centerpoint kernel size}
For 2D object detection models a kernel size of $3$ is often used, but since the 3D adaptation problem is different in performing adaptation of \textit{BEV}-features instead of image features, different kernel sizes were tested. Kernel sizes of $1$, $3$, and $5$ were tested for the discriminator layers in Centerpoint~\cite{yin2021center}. Results are found in \Cref{tab:params-kernelsize}.

\begin{table}[h]
    \centering
    \caption{Different kernel sizes for Centerpoint~\cite{yin2021center}. For the unsupervised task of \textit{VLD-64} $\rightarrow$ \textit{VLD-16}.}
    \label{tab:params-kernelsize}
    \resizebox{\columnwidth}{!}{%
    \begin{tabular}{c|c|ccc|c}
        & \multicolumn{5}{c}{Centerpoint} \\
        \cline{2-6}
        &  & \multicolumn{3}{|c|}{Kernel Size} &   \\
        & Source-Only & $1$ & $3$ & $5$ & Oracle  \\
        \hline
        $AP_{3D, V}$  & 32.30 & 43.90 & \textbf{46.99 }& 46.55 & 53.54 \\    
        $AP_{3D, P}$  & 11.63 & 32.92 & \textbf{38.22} & 37.52 & 47.78 \\ 
        $AP_{3D, C}$  & 23.06 & 46.20 & 45.56 & \textbf{46.62} & 53.11 \\ 
    \end{tabular}}
    \label{tbl:params-kernelsize}
\end{table}


In \Cref{tbl:params-kernelsize} it can be seen that the results achieved using a kernel size of $5$ and $3$ are almost the same. Larger kernel sizes however result in higher GPU usage. While using models with a kernel size of $5$ the GPU usage during training with a batch size of $8$ resulted in a GPU usage just slightly under the available $24$\,GB of the GPU used. Therefore, a kernel size of $3$ was chosen to avoid memory overloading. The Centerpoint~\cite{yin2021center} implementation uses the standard voxel size of $0.1$\,m in $x$ and $y$, and $0.15$\,m in $z$.

\section{Random Object Scaling}

Random Object Scaling (ROS) applies random scaling factors to ground truth bounding boxes and their corresponding points. Each object point in ego-vehicle frame $(p_i^x,p_i^y,p_i^z)_{\mathrm{ego}}$ is transformed to local object coordinates

\begin{equation}
    \begin{array}{cc}
         (p_i^l,p_i^w,p_i^h)_{\mathrm{object}} = ((p_i^x,p_i^y,p_i^z)_{\mathrm{ego}} -  \\
         ~~~~~~~~~~~~~~(c_x,c_y,c_z)_{\mathrm{object}}) \times R_{\mathrm{object}}
    \end{array}
\end{equation}

\noindent
where $(c_x,c_y,c_z)_{\mathrm{object}}$ is object center coordinates and $R_{\mathrm{object}}$ is the rotation matrix between the ego-coordinates and object-coordinates. Each object point is then scaled by a random scaling factor $r$ drawn from a uniform random distribution:

\begin{equation}
    \begin{array}{cc}
         (p_i^l,p_i^w,p_i^h)_{\mathrm{object},\mathrm{scaled}} = r \cdot (p_i^l,p_i^w,p_i^h)_{\mathrm{object}} \;  \\
         r \in U(r_{\mathrm{min}},r_{\mathrm{max}}).
    \end{array}
\end{equation}

Afterwards, each object is transformed back into the ego-vehicle frame. The length $l$, width $w$, and height $h$ of each bounding box are also scaled accordingly with $r$.

Following previous works~\cite{wang2020train,saltori2020sf,yang2021st3d} experiments are performed by including object scaling in the source-data to account for different vehicle sizes. As examined in Section 4.1 there is a large difference between the vehicle sizes in LiDAR-CS, corresponding to large vehicle sizes typically found in the USA, and the smaller vehicles in Europe encountered by the robot, which encounters the smaller vehicles in Europe. Specifically, ROS from ST3D~\cite{yang2021st3d} is utilized where the objects and corresponding bounding boxes are scaled according to uniform noise in a chosen scaling interval. While previous UDA methods on LiDAR-based 3D object detection often apply domain adaptation only to a single object category, we consider it a multiclass problem. Therefore, ROS is used for all three classes (Vehicle, Pedestrian, and Cyclist) with different scaling intervals, shown in~\cref{tab:ros_scale}. The results of different scaling intervals are shown in \cref{tab:ros_performance}. We can see that the best setting is $\Delta_{x,y,z} \in \{0.80,1.20\}$ for vehicles and $\Delta_{x,y,z} \in \{0.9,1.1\}$ for pedestrians and cyclists, thus, we chose this for all of our experiments.  


\begin{table}[h]
    \centering
    \caption{Tested multi-class scaling intervals for ROS.}
    \label{tab:ros_scale}
    \resizebox{\columnwidth}{!}{
    \begin{tabular}{c|ccc}
        Scaling  & Vehicles & Pedestrians & Cyclists \\
        Settings&  &  &  \\
        \hline
        1 & $U(0.8,1.0)$ & $U(0.8,1.0))$ & $U(0.8,1.0)$ \\
        2 & $U(0.9,1.1)$ & $U(0.8,1.2)$ & $U(0.8,1.2)$ \\
        3 & $U(0.8,1.2)$ & $U(0.9,1.1)$ & $U(0.9,1.1)$ \\
    \end{tabular}}
    
    \label{tbl:laura-scaling-factors}
\end{table}

\begin{table}[h!]
    \centering
    \caption{Random Object Scaling using different scaling ranges tested on \textit{CS-16} $\rightarrow$ \textit{Robot}.} 
    \label{tab:ros_performance}
    \resizebox{\columnwidth}{!}{%
    \begin{tabular}{c|cccc}
         &Source & $v_{0.8,1.0}$ & $v_{0.8,1.2}$ & $v_{0.9,1.1}$  \\
        & Only & $pc_{0.8,1.0}$ & $pc_{.8,1.2}$ & $pc_{0.9,1.1}$ \\
       
        \hline
        $AP_{3D, V}$    & 28.96  & 46.17  & \textbf{50.83} & 44.66 \\    
        $AP_{3D, P}$    & 11.26 & 37.39  & \textbf{40.36} & 39.36 \\ 
        $AP_{3D, C}$    & 4.99 & \textbf{31.22} & 30.23 & 25.76 \\ 
        $mAP_{3D}$    & 15.07 & 38.73& \textbf{41.47} & 39.93 \\ 

    \end{tabular}}
    \label{tbl:ROS-ranges}
\end{table}

Regarding downsampling and ROS, we used downsampled point clouds for training \textit{teacher models} for all the methods presented in Table I of the main paper in order to bridge the initial gap. We also applied uniform random object scaling (ROSC) for those methods. However, while other methods used uniform object scaling, our method employed the nonuniform ROS that we developed. In~\cref{tab:ros1,tab:ros2}, we present our method with different options: no random object scaling and no downsampling (\textbf{UADA3D -ROS/-DS}), no random object scaling with downsampling (\textbf{UADA3D -ROS/+DS}), with our random object scaling, no downsampling (\textbf{UADA3D -ROS/+DS}), with uniform random object scaling and downsampling (\textbf{UADA3D +ROSC/+DS}) and finally with our \textit{original method}: random object scaling, and downsampling (\textbf{UADA3D +ROS/+DS}).

\begin{table}[htbp]
\centering
\caption{Performance Comparison of UDA on IA-SSD with different ROS approaches on Waymo to nuScenes (W-N) and LiDAR-CS 32 to Laura Robot data (CS32-R)}
\label{tab:ros1}
\resizebox{\linewidth}{!}{
\begin{tabular}{clccc}
\toprule
& \textbf{Method} & \textbf{mAP3D/BEV} & \textbf{Change} & \textbf{Closed Gap \%} \\ \midrule\midrule
\multirow{7}{*}{W-N}& \textbf{Source Only}    & 1.5 / 4.50     & -       & -       \\
& \textbf{UADA3D -ROS/-DS}           & 13.71 / 14.42 & 12.21 / 9.92   & 27.31 / 23.05 \\
& \textbf{UADA3D -ROS/+DS}           & 14.15 / 15.07 & 12.65 / 10.57   & 28.29 / 24.56 \\
& \textbf{UADA3D +ROS/-DS}           & 17.10 / 17.44  & 15.60 / 12.94   & 34.90 / 30.07 \\
& \textbf{UADA3D +ROSC/+DS}          & 18.31 / 17.92  & 16.81 / 13.42   & 37.60 / 31.19 \\
& \textbf{UADA3D +ROS/+DS}           & \textbf{18.33 }/ \textbf{18.55}  & 16.83 / 14.05  & 37.66 / 32.66 \\
& \textbf{Oracle}         & 46.20 / 47.53  & -   & - \\ \bottomrule

\multirow{7}{*}{CS32-R}& \textbf{Source Only}    & 8.33 / 15.93     & -       & -      \\
& \textbf{UADA3D -ROS/-DS}           & 18.42 / 25.73 & 10.09 / 9.80  & 28.20 / 27.09 \\
& \textbf{UADA3D -ROS/+DS}           & 20.04 / 27.98   & 11.74 / 12.05   & 32.78 / 33.31 \\
& \textbf{UADA3D +ROS/-DS}           & 20.91 / 28.58   & 12.58 / 12.65   & 35.16 / 34.97 \\
& \textbf{UADA3D +ROSC/+DS}           & 21.36 / 30.22  & 13.03 / 14.29   & 36.42 / 39.51 \\
& \textbf{UADA3D+ROS/+DS}           & \textbf{22.15}/ \textbf{31.90}   & 13.82 / 15.97   & 38.62 / 44.14 \\

& \textbf{Oracle}         & 44.11 / 52.10   & -   &  - \\ \bottomrule
\end{tabular}
}
\end{table}

\begin{table}[htbp]
\centering
\caption{Performance Comparison of UDA on Centerpoint with different ROS approaches on Waymo to nuScenes (W-N) and LiDAR-CS 32 to Laura Robot data (CS32-R).}
\label{tab:ros2}
\resizebox{\linewidth}{!}{
\begin{tabular}{clccc}
\toprule
& \textbf{Method} & \textbf{mAP3D/BEV} & \textbf{Change} & \textbf{Closed Gap \%} \\ \midrule\midrule
\multirow{7}{*}{W-K} & \textbf{Source Only}    & 15.96 / 17.87   & -          & -           \\
& \textbf{UADA3D -ROS/-DS}           & 18.31 / 21.93   & 2.35 / 4.06      & 6.37 / 11.01  \\
& \textbf{UADA3D -ROS/+DS}           & 19.12 / 23.65   & 3.16 / 5.78       & 8.57 / 15.66   \\
& \textbf{UADA3D +ROS/-DS}           & 23.86 / 28.17   & 7.90 / 10.30      & 21.42 / 27.91   \\
& \textbf{UADA3D +ROSC/+DS}           & 26.09 / \textbf{31.91}   & 10.13 / 12.04      & 27.47 / 32.63   \\
& \textbf{UADA3D +ROS/+DS}           & \textbf{26.89} / 30.67   & 10.94 / 12.80    & 29.65 / 34.68   \\

& \textbf{Oracle}         & 48.39 / 59.54 & -     & -     \\ \midrule

\multirow{7}{*}{CS32-R}  & \textbf{Source Only}    & 3.78 / 6.00     & -          & -           \\

& \textbf{UADA3D -ROS/-DS}           & 23.11 / 28.62  & 19.33 / 22.62    &   43.33 / 42.25 \\
& \textbf{UADA3D -ROS/+DS}           & 24.09 / 29.71   &  20.31 / 23.71   & 45.52 / 44.28  \\
& \textbf{UADA3D +ROS/-DS}           & 28.35 / 38.14   & 24.57 / 32.14     & 55.08 / 60.03  \\
& \textbf{UADA3D +ROSC/+DS}           & 30.69 / \textbf{43.38}   & 26.91 / 35.38     &  60.32 / 66.08  \\
& \textbf{UADA3D +ROS/+DS}           & \textbf{31.54}/ 42.82   & 27.76 / 36.83    & 62.23 / 68.78   \\

& \textbf{Oracle}         & 48.39 / 59.54   & -     & -       \\ \bottomrule
\end{tabular}
}
\end{table}

The results presented in~\cref{tab:ros1,tab:ros2}, highlight the effectiveness of UADA3D with Rescaling and Downsampling (ROS and DS) in improving performance. On both the IA-SSD and CenterPoint, UADA3D with ROS and DS (UADA3D +ROS/+DS) consistently achieves the highest mAP3D and BEV scores compared to other configurations and methods. Specifically, regardless which setting we use UADA3D significantly outperforms \textit{Source Only} models. These improvements highlight UADA3D effectiveness in adaptation, showing that while ROS and DS provide valuable performance boosts, they are secondary to the benefits delivered by the UADA3D approach itself. The fact that UADA3D with or without ROS and DS outperforms other methods, including those using ROSC and downsampling, further reinforces that the core advancements are attributable to the method's inherent capabilities rather than the additional features.

\subsubsection{GRL parameter}
\begin{figure}[ht!]
    \centering
    \includegraphics[width=\columnwidth]{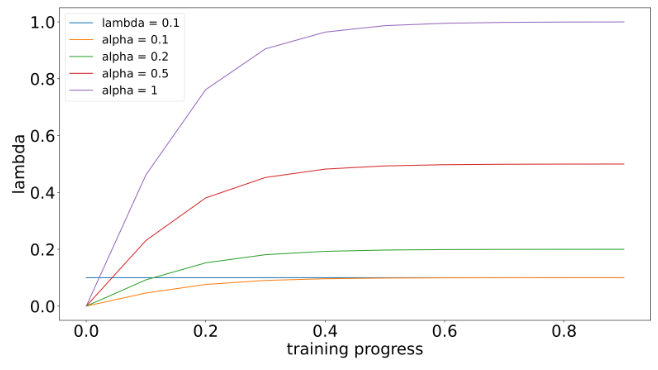}
    \caption{Functions for the gradient reversal coefficient $\lambda$.}
    \label{fig:grl-curves}
\end{figure}

In ablation studies we tested two different strategies for \textit{GRL}-coefficient $\lambda$ on UADA3D and UADA3D\textsubscript{$\mathcal{L}m$}. Firstly, a stationary $\lambda=0.1$ was tested following the setting used for most adversarial UDA strategies in 2D object detection~\cite{chen2018domain,saito2019strong}. Secondly, we follow other approaches~\cite{ganin2015unsupervised,li2023domain} where $\lambda$ was increased over the training according to:

\begin{equation} \label{eq:params-grl}
    \lambda = \alpha (\frac{2}{1 + exp(-\gamma p)} - 1)
\end{equation}

where $\alpha \in [0,1]$ is a scaling factor that determines the final $\lambda$, $\gamma=10$ and $p$ is the training progress from start $0$ to finish $1$. $\alpha$-values of $1, 0.5, 0.2, 0.1$ were tested. The $\lambda$ parameter for different values of $\alpha$ throughout training process is illustrated in \cref{fig:grl-curves}. Numerical results regarding the influence of $\lambda$ on the model performance are available in Ablation Studies. 

We also tried lower values but runs using $\lambda = 0.1-0.15$ performed the best. Additionally, we considered using an $\alpha$ value between $0.05 - 0.15$, as smaller dynamic $\lambda$ seemed to work better for pedestrians. However, while this showed promising results for pedestrians, it, as in Table IV, performed worse for vehicles and cyclists.

In \cref{tab:grlcnst} we observe that values between 0.09 and 0.2 could potentially yield better results, but $\lambda = 0.1$ appears to work best overall for $mAP_{3D}$ across the three classes. In the same table, we can see that as $\lambda$ decreases, our model gradually converges towards the source-only model, which makes sense since a lower $\lambda$ weakens the learning signal from the discriminator.

\begin{table}[ht]
\centering
\caption{Gradient reversal layer coefficient $\lambda$ tests of CS64-CS16 adaptation on IA-SSD.}
\label{tab:grlcnst}
\begin{tabular}{c|c|c|c|>{\columncolor[gray]{0.8}}c}
$\lambda$ & $AP_{3D,V}$ & $AP_{3D,P}$ & $AP_{3D,C}$ & $mAP_{3D}$ \\ \midrule
Source & 29.97 & 11.02 & 19.35 & 20.78 \\ \hline
0.01 & 31.13 & 16.77 & 21.43 & 23.11 \\ \hline
0.02 & 31.89 & 15.18 & 22.80 & 23.29 \\ \hline
0.05 & 38.13 & 23.41 & 29.29 & 30.94 \\ \hline
0.07 & 39.66 & 23.14 & 34.77 & 32.52 \\ \hline
0.09 & 41.06 & \textbf{26.09} & 37.95 & 35.03 \\ \hline
0.1 & 42.96 & 25.48 & \textbf{38.12} & \textbf{35.52} \\ \hline
0.2 & \textbf{43.24} & 24.86 & 34.24 & 34.11 \\ \hline
0.3 & 42.63 & 24.69 & 32.81 & 33.38 \\ \hline
0.4 & 40.11 & 24.07 & 32.05 & 32.08 \\ \hline
0.5 & 39.60 & 23.63 & 31.45 & 31.56 \\ \bottomrule
Oracle & 58.62 & 35.57 & 49.27 & 47.82 \\ 
\end{tabular}
\end{table}

\section{Datasets}

We acknowledge the importance of this information and we will added the note on that to the paper and we will include the detailed table to the supplementary materials. In UADA3D training, we use \textit{train samples} from source and target datasets. In each batch we randomly number of samples from source and target data. Note, the number of samples from the target and source data differ in each batch. The probability of drawing the sample from the data is proportional to its size since we see each sample once per epoch and we go over the whole training split once each epoch. For other methods and oracle models we use train split for training the oracle/teacher model and then train split of the target data for adaptation.  

Since the Waymo and nuScenes are larger, we considered reducing them to 7000 train samples, however, that would create unfair comparison since the oracle models trained on those datasets would be trained on much smaller number of samples than other projects reported and would create unfair comparison, thus we used the original split. 

\begin{table}[h!]
\centering
\resizebox{\columnwidth}{!}{
\begin{tabular}{c|c|c}
\textbf{Dataset} & \textbf{Number of Train Samples} & \textbf{Number of Test Samples} \\ \hline
\textbf{KITTI}   & \textasciitilde 7480           & \textasciitilde 7520           \\ \hline
\textbf{nuScenes} & 28130                         & 6000                           \\ \hline
\textbf{Waymo}   & \textasciitilde 158100         & \textasciitilde 40000          \\ \hline
\textbf{Laura}   & \textasciitilde 7000        & \textasciitilde 3000          \\ \hline
\textbf{LiDAR-CS$^*$}   & \textasciitilde 7000        & \textasciitilde 7000          \\ 
\end{tabular}
}
\caption{Train-test split for the datasets. *Note, the each LiDAR-CS \textit{subdataset} (CS64, CS32, CS16) contains this number of samples.}
\end{table}

In \cref{tab:lidars} we compare different LiDARs's properties, their number of points per scan as well as height above ground in the different datasets. The variety of scenarios we tested clearly shows robustness of our and different UDA approaches, specifically when applied towards sparser data with large domain gaps. Comprehensive data analysis is available in the main paper (Section 4).

\begin{table}[h!]
    \centering
    \caption[\textit{LiDAR-CS} dataset details]{LiDAR sensors details.}
    \label{tab:lidars}
    
    \resizebox{\columnwidth}{!}{
    \begin{tabular}{c|lrr}
         Sensor & FOV          & $n_{points}/$ scan &  LiDAR Height \\
         \hline
         HLD-64 (Kitti)  & $26.9^\circ$ & $\approx 118000$ & $\approx1.6$\,m \\
         VLD-64 (CS64)& $26.9^\circ$ & $\approx 100000$   & $\approx2$\,m \\
         HLD-64 (Waymo) & $26.9^\circ$ & $\approx 144512$   & $\approx1.6$\,m \\
        VLD-32 (CS32) & $40^\circ$ & $\approx 63900$ &$2$\,m\\ 
        VLD-32 (nuScenes) & $40^\circ$ & $\approx 34688$ &$1.6$\,m\\ 
        VLD-16 (CS16) & $30^\circ$   & $\approx 22000$  & $2$\,m\\
        VLP-16 (robot) & $30^\circ$   & $\approx 22000$  & $0.6$\,m\\

    \end{tabular}}
    \label{tab:details-CS}
\end{table}

\section{Sparse vs. depth dense perception}
 
When comparing sparse and dense perception in the context of LiDAR for robotics or autonomous vehicles, the core distinction revolves around the density of the point cloud data provided by the LiDAR sensor. This density impacts perception algorithms, computational requirements, and overall accuracy. Dense perception is crucial in applications where high accuracy and detail are necessary, such as autonomous vehicles navigating complex urban environments. Dense LiDAR is also essential for tasks requiring high-precision, such as mapping and localization in dynamic environments. Sparse perception, on the other hand, may be more suitable for simpler robotic tasks such as indoor navigation or agricultural robotics, where fine details are less critical, and real-time performance is essential. 

Dense LiDAR, such as a 128-layer or 64-layer sensor, captures a higher number of laser returns per rotation, providing a more detailed and denser point cloud.

\textbf{Advantages:}
\begin{itemize}
    \item \textbf{High Resolution:} Dense perception offers fine-grained representations of the environment, capturing more details about objects, terrain, and small obstacles.
    \item \textbf{Improved Object Detection:} Higher resolution results in better object detection and classification, especially for smaller or distant objects.
    \item \textbf{Enhanced Depth Accuracy:} Depth estimation becomes more reliable due to the larger volume of data, improving the accuracy of 3D environmental representation.
    \item \textbf{Reduced Data Gaps:} Dense LiDAR results in fewer blind spots, benefiting algorithms reliant on complete coverage, such as voxel-based perception.
\end{itemize}

\textbf{Challenges:}
\begin{itemize}
    \item \textbf{High Computational Load:} Dense point clouds require significantly more computational resources, including memory, processing time, and communication bandwidth.
    \item \textbf{Data Handling Overhead:} Transmitting and storing large point clouds requires optimized data handling, posing challenges for real-time systems.
    \item \textbf{Cost:} Dense LiDAR systems are generally more expensive and require robust hardware and cooling systems.
\end{itemize}

In contrast, Sparse Perception (Sparse LiDAR: 32-1 Layers) generates fewer data points, creating a sparser representation of the environment.

\textbf{Advantages:}
\begin{itemize}
    \item \textbf{Lower Computational Requirements:} Sparse LiDAR generates less data, reducing computational, memory, and bandwidth demands, making it easier to achieve real-time performance.
    \item \textbf{Lower Cost:} Sparse LiDAR sensors are more affordable and suitable for low-budget platforms.
    \item \textbf{Sufficient for Certain Tasks:} Sparse perception can suffice for simpler tasks such as obstacle avoidance and rough terrain mapping.
\end{itemize}

\textbf{Challenges:}
\begin{itemize}
    \item \textbf{Reduced Resolution:} The lower resolution impacts the detection of smaller or distant objects, and there may be more blind spots in the environment.
    \item \textbf{Less Reliable Depth Estimation:} Sparse point clouds result in noisier depth estimation, reducing the accuracy of 3D reconstructions.
    \item \textbf{Need for Interpolation:} Sparse data often requires additional algorithms to interpolate or densify the point cloud, which can introduce errors or computational overhead.
\end{itemize}


\clearpage
\noindent

\end{document}